\documentclass{article}
\usepackage{silence}
\usepackage{amsmath}
\usepackage{amssymb}
\usepackage[mathscr]{euscript}
\usepackage{txfonts}
\usepackage{stmaryrd}

\numberwithin{equation}{section}

\usepackage[table,dvipsnames]{xcolor}
\definecolor{shadecolor}{gray}{0.9}

\usepackage{amsthm}
\theoremstyle{definition}

\usepackage[final,expansion=alltext]{microtype}
\usepackage[english]{babel}
\usepackage[parfill]{parskip}
\usepackage{afterpage}
\usepackage{framed}
\usepackage{verbatim}
\usepackage{setspace}
\usepackage{bbold}
\usepackage{wrapfig}

{\endMakeFramed}

\usepackage{ragged2e}

\newcounter{parcount}

\usepackage[symbol]{footmisc}
\renewcommand{\thefootnote}{\arabic{footnote}}

\usepackage[inline]{enumitem}

\DeclareMathAlphabet{\pazocal}{OMS}{zplm}{m}{n}

\usepackage{graphicx}
\usepackage[labelfont={it}]{caption}
\usepackage[format=hang]{subcaption}
\usepackage{wrapfig}

\usepackage{chngcntr}

\usepackage{booktabs}
\usepackage{longtable}
\usepackage{hhline}

\usepackage{listings}
\usepackage{fancyvrb}
\fvset{fontsize=\normalsize}

\usepackage[colorlinks,linktoc=all]{hyperref}
\usepackage[all]{hypcap}
\hypersetup{citecolor=MidnightBlue}
\hypersetup{linkcolor=black}
\hypersetup{urlcolor=MidnightBlue}

\usepackage[acronym,smallcaps,nowarn]{glossaries}[=v4.46]

\lstdefinestyle{mystyle}{
    commentstyle=\color{OliveGreen},
    keywordstyle=\color{BurntOrange},
    numberstyle=\tiny\color{black!60},
    stringstyle=\color{MidnightBlue},
    basicstyle=\ttfamily,
    breakatwhitespace=false,
    breaklines=true,
    captionpos=b,
    keepspaces=true,
    numbers=left,
    numbersep=5pt,
    showspaces=false,
    showstringspaces=false,
    showtabs=false,
    tabsize=2
}
\usepackage{soul}

\renewcommand{\mid}{~\vert~}

\let\HyAddContentsLine\addcontentsline
\usepackage[final]{colm2026_conference}
\let\addcontentsline\HyAddContentsLine

\usepackage{microtype}
\usepackage{hyperref}
\usepackage{url}
\usepackage{booktabs}
\usepackage{cleveref}
\usepackage{nicefrac}
\usepackage{empheq}

\usepackage{lineno}

\definecolor{darkblue}{rgb}{0, 0, 0.5}
\hypersetup{colorlinks=true, citecolor=darkblue, linkcolor=darkblue, urlcolor=darkblue}

\newcommand{\softmax}{softmax }
\newcommand{\Softmax}{Softmax }

\newcommand{\method}{SwiLA }
\newcommand{\pmethod}{SwiLA}
\newcommand{\gi}{j}
\newcommand{\GI}{J}
\newcommand{\pe}{\delta}
\newcommand{\SW}{\theta}

\crefname{appendix}{appendix}{appendices}
\Crefname{appendix}{Appendix}{Appendices}

\title{Switching Linear Attention}

\author{%
  Hyun Dong Lee$^*$ \\
  Stanford University
  \And
  Xavier Gonzalez \\
  Stanford University$^\dagger$
  \And
  Nicolas Zucchet \\
  Stanford University
  \AND
  E. Kelly Buchanan \\
  Stanford University
  \And
  Emily B. Fox \\
  Stanford University
  \And
  Scott W. Linderman \\
  Stanford University
}

\begin{document}

\ifcolmsubmission
\linenumbers
\fi

\maketitle

{\let\thefootnote\relax
 \makeatletter\let\@footnotetext\H@@footnotetext\makeatother
 \footnotetext{$^*$Correspondence to \texttt{hdlee@cs.stanford.edu}.}
 \footnotetext{$^\dagger$Now at Unconventional AI. Work done at Stanford University.}
 \footnotetext{$^\ddagger$Code is available at \href{https://github.com/lindermanlab/switching-linear-attention}{\texttt{switching-linear-attention}}.}
}

\begin{abstract}
  Designing expressive sequence layers with efficient inference remains a central challenge in modern machine learning. 
Standard \softmax attention achieves excellent sequence modeling performance through rich nonlinear token interactions, but it requires a key--value cache that grows linearly with sequence length, limiting its scalability. 
Linear attention enables efficient recurrent computation with a constant memory footprint, yet its reduced expressivity often yields inferior modeling performance.
We introduce \emph{Switching Linear Attention} (\pmethod), a novel sequence layer that bridges this gap by enhancing representational capacity while retaining the fixed-size recurrent state of linear attention.
We derive the \method recurrence from the test-time regression framework, casting the state update rule as online expectation--maximization in a mixture of linear regressions model.
At test time, each output dimension dynamically selects among multiple linear attention components based on the input.
Across associative recall, in-context language learning, and language modeling benchmarks, \method shows strong performance and narrows the gap to softmax attention, even surpassing it in several settings.
\end{abstract}

\section{Introduction}

Attention mechanisms \citep{vaswani2017attention} have emerged as the cornerstone of modern sequence modeling, powering state-of-the-art advances in language understanding~\citep{brown2020language}, visual
recognition~\citep{dosovitskiy2020image}, and multimodal vision-language
reasoning~\citep{alayrac2022flamingo}. 
The key innovation is dynamic information routing based on input-dependent similarity, enabling models to selectively retrieve relevant context from across a sequence. 
A useful perspective on this retrieval capability is that softmax attention implicitly fits a nonlinear mapping between inputs and outputs at inference time, a form of \emph{test-time regression} (TTR) that underlies its expressive in-context learning (ICL)~\citep{garg2022icl, akyurek2022what, vonoswald2023transformers, wang2025test}.

However, this expressivity comes at a high computational cost. 
The key-value (KV) cache required for autoregressive generation grows linearly with sequence length, which becomes prohibitive for long-context applications.
Linear attention~\citep{katharopoulos2020transformers} addresses this bottleneck by replacing softmax with a kernel feature map, reformulating attention as a linear recurrence with a state of size $O(D^2)$, where $D$ is the embedding dimension. 
However, this compression fundamentally limits associative recall: at each timestep, linear attention retrieves values through a single linear mapping of the query, and cannot capture the diverse, context-dependent retrieval patterns that softmax attention achieves through its nonlinear mapping.
Empirically, linear attention consistently underperforms on tasks requiring precise memory retrieval~\citep{zoology2024, arora2024simple}.

To bridge this gap, we introduce \emph{Switching Linear Attention} (\pmethod), a novel sequence layer that leverages the TTR framework of~\citet{wang2025test}, extending it to a mixture of linear regressors$^\ddagger$.
\method divides its recurrent state into $\GI$ mixture components, each maintaining its own linear regressor of size $O(D^2)$ that specializes in different retrieval patterns. 
At each timestep, each of the $D$ output dimensions selects a mixture component based on both a learned gating prior and the prediction errors of the current state, yielding a number of mixtures that scales exponentially as $O(\GI^D)$, despite having a state size that grows only polynomially as $O(\GI D^2)$.
We derive the update rule from online expectation maximization~\citep{cappe2009line}: the E-step computes posterior responsibilities based on prediction errors, and the M-step updates each component's regression weights.
During in-context learning, \method learns an expressive \emph{piecewise} linear mapping from inputs to outputs, while retaining, as in linear attention, a fixed-size recurrent state whose memory footprint is independent of sequence length.
With experiments on associative recall, in-context language learning, and language modeling, we find that \method performs strongly, narrowing the gap to softmax attention and even surpassing it in several settings.

\section{Background on Test-Time Regression}
\label{sec:background}

A central capability of effective sequence models is \emph{in-context associative recall}: given a query, retrieve relevant information from previously observed tokens.
\citet{wang2025test} formalize in-context associative recall as a two-step process of memorization and retrieval. They show that the memorization step can be cast as an online regression problem, which they call \textit{test-time regression} (TTR).

\paragraph{Memorization as regression.}
Given a sequence of KV pairs $(k_1, v_1), \ldots, (k_T, v_T)$ with $k_t \in \mathbb{R}^D$ and $v_t \in \mathbb{R}^D$, an associative memory is a function $m_t : \mathbb{R}^D \to \mathbb{R}^D$ such that $m_t(k_i) \approx v_i$ for $i = 1, \ldots, t$. 
Finding such a map reduces to solving a regression problem:
\begin{equation}
    m_t \approx \arg\min_{m \in \mathcal{M}} \sum_{i=1}^{t} \| v_i - m(k_i) \|^2,
    \label{eq:ttr-memorize}
\end{equation}
where $\mathcal{M}$ is a chosen function class.

\paragraph{Retrieval as function evaluation.}
Once the regressor $m_t$ has been fit, memory retrieval is simply function application: given a query $q_t$, the output is
\begin{equation}
    o_t = m_t(q_t).
    \label{eq:ttr-retrieve}
\end{equation}
A sequence layer that solves~\cref{eq:ttr-memorize} and applies~\cref{eq:ttr-retrieve} at each timestep performs associative recall in its forward pass. 
This framework is called test-time regression (TTR): the regressor $m_t$ is learned in-context from the input tokens and the structure of the recurrence. Commonly, the query $q_t$, key $k_t$, and value $v_t$ are all projections of some input $x_t$.

\paragraph{Design space.}
In TTR, a sequence layer is specified by the choice of the function class $\mathcal{M}$ and the optimization algorithm for~\cref{eq:ttr-memorize}.
These choices, along with the particular keys and values shown at test-time, ultimately determine the regressor $m_t$.
In the parametric case, $m_t$ corresponds to the fixed-size weights of a function approximator, updated online as each new pair $(k_t, v_t)$ arrives.
In the nonparametric case, we recover kernel regression.
Importantly, $m_t$ is not learned during training; it is learned in-context from the input tokens at inference time, by the very structure of the recurrence.
Existing sequence layers arise as special cases.
Linear attention~\citep{katharopoulos2020transformers} uses a linear model for $m_t$, fit with a whitened design matrix approximation.
DeltaNet~\citep{schlag2021linear, yang2024parallelizing} also employs a linear model but uses SGD as its optimization algorithm.
Softmax attention fits a nonparametric kernel smoother.
We refer the reader to~\citet{wang2025test} for a comprehensive taxonomy and to Appendix~\ref{app:ttr_design_space} for further details.

\subsection{Limitations motivating Switching Linear Attention}
\begin{figure}[!t]
    \centering
    \includegraphics[width=\textwidth]{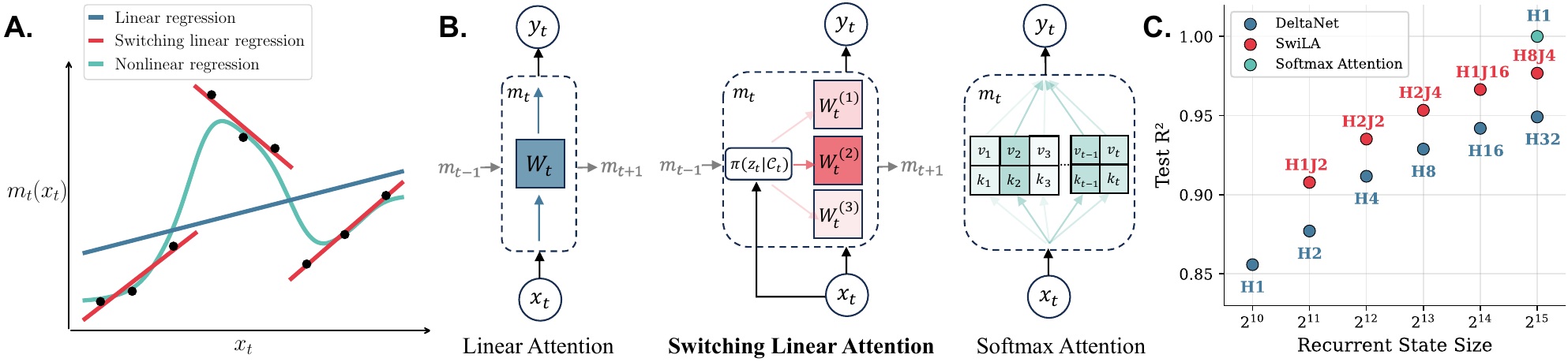}
    \caption{\textbf{Switching Linear Attention.} 
    \textbf{(A)}: When the input--output relationship varies with context, piecewise ("switching") regression is more expressive than linear regression.
    \textbf{(B)}: 
    \method increases expressivity toward that of softmax attention while retaining, as in linear attention, a fixed-size recurrent state whose memory footprint is independent of sequence length.
    \pmethod's state is a mixture of linear regressors.
    At each step, \method selects which regressor to use based on the current input and previous state.
    \textbf{(C)}: Test $R^2$ vs.\ state size for \method and DeltaNet on a synthetic softmax attention regression task, with \softmax attention for reference.
    Labels mark the best \method configuration (\texttt{H}$=$heads, \texttt{\GI}$=$mixtures). 
    At every budget, \method better approximates softmax attention.}
    \label{fig:sla_schematic}
\end{figure}

TTR reveals a fundamental tension. 
Linear attention and its variants compress all history into a single fixed-size state, bounding the amount of context that can be stored.
At each timestep, retrieval is a single linear mapping of the query, which cannot capture the diverse, context-dependent retrieval patterns that \softmax attention achieves through its nonlinear mapping. 
Softmax attention avoids both limitations by storing all KV pairs and retrieving via a nonlinear kernel smoother, but at the cost of a KV cache that grows with sequence length.
\method defines a piecewise linear map that can approximate arbitrary nonlinear functions (\cref{fig:sla_schematic}A) by replacing the single linear regressor with a \emph{mixture of linear regressors} (\cref{fig:sla_schematic}B).
Thus, \method addresses the retrieval limitation of linear attention while retaining a fixed-size recurrent state whose memory footprint is independent of sequence length.

\section{Switching Linear Attention}
\label{sec:sla}

We introduce \textit{Switching Linear Attention} (\pmethod), a sequence layer based on mixtures of linear regressions. 
\method retains the fixed-size recurrent state of linear attention, while improving its retrieval expressivity through a nonlinear mapping from query to output in \cref{eq:ttr-retrieve}, as in softmax attention.
\method is a form of TTR with,
\begin{itemize}
    \item Regression model: mixture of linear regressors \citep{pearson1894mixtures, de1989mixtures}
    \item Optimization algorithm: online expectation--maximization (EM) \citep{dempster1977maximum, cappe2009line}.
\end{itemize}

For the rest of the paper, we consider a sequence of keys $k_t$, values $v_t$, and queries $q_t$ in $\mathbb{R}^D$, projected from the inputs $x_t \in \mathbb{R}^{d_{\text{in}}}$.
Further technical details are given in \Cref{app:tech_details}.

\subsection{Regression Model: Mixture of Linear Regressors}\label{ssc:sla_mod}

Mixture modeling is a foundational approach for modeling functions that vary based on group membership \citep{bishop2006textbook}.
Consider a dataset $\{(k_t, v_t)\}_{t=1}^T$ consisting of key--value pairs with
$k_t \in \mathbb{R}^D$ and $v_t \in \mathbb{R}^D$, assumed to be generated from a mixture of $\GI$ linear Gaussian regression models.
To improve expressivity in high-dimensional settings, we allow group membership to vary across the output dimension $d$.

Specifically, let $z_{td} \in \{1,\ldots,\GI\}$ denote 
the group membership of data point $t$ and output dimension $d$.
Then, given a key $k_t$, the value $v_{td}$ is modeled as
\begin{align}\label{eq:molr_gen_model}
    p(z_{td} \mid k_t) = \mathrm{Cat}\left(z_{td} \mid \pi_{jd}(k_t)\right), \qquad 
    p(v_{td} \mid z_{td}=\gi, k_t) = \mathcal{N}\!\left(v_{td} \mid w_{\gi d}^\top k_t,\; 1 \right).
\end{align}
Here, $\pi_{\gi d}(k_t)$ denotes the prior probabilities of the mixture components. The regression weights, $w_{\gi d} \in \mathbb{R}^D$, are specific to mixture component $\gi$ and output dimension $d$.
We denote $W \in \mathbb{R}^{\GI \times D \times D}$ as the collection of all weights.
Since each output dimension independently selects its own mixture component, the effective number of mixtures grows exponentially as $\GI^D$, whereas the number of weights scales polynomially as $\GI D^2$.

Under the probabilistic model in \cref{eq:molr_gen_model}, the marginal log-likelihood for timestep~$t$ and output dimension~$d$ is
\begin{align}
    \ell_{td} = \log p(v_{td} \mid k_t) = \log \sum_{\gi=1}^{\GI} \pi_{\gi d}(k_t)\,\exp\!\left(-\tfrac{1}{2} \bigl(v_{td} - w_{\gi d}^\top k_t\bigr)^2\right) + c. \label{eq:loglik}
\end{align}
Complete modeling details are given in \Cref{app:sla_details}.

\subsection{Optimization Algorithm: Online EM}\label{ssc:sla_alg}

The standard approach for fitting mixtures of linear regressors \eqref{eq:molr_gen_model} is the EM algorithm. 
In the TTR setting, the weights $W$ are estimated online: at each timestep $t$, the current estimates $W_{t-1}$ are updated as the new pair $(k_t, v_t)$ arrives.
Our choice of online EM as the optimizer is what gives rise to the \method recurrence.
It amounts to a one-step stochastic gradient update, where we ascend on the marginal log-likelihood \eqref{eq:loglik} with respect to the "fast-weights" $W$. 
In particular, \textbf{the \method recurrence} for all mixture components $\gi$ and output dimensions $d$ is
\begin{align}
    w_{t\gi d} & =  w_{t-1,\gi d} + \beta_{t\gi d} \nabla_{w_{t-1, \gi d}} \ell_{td} \nonumber \\
    & = w_{t-1,\gi d} + \beta_{t \gi d}\, r_{t\gi d}\,   \delta_{t\gi d}\,  k_t. \label{eq:recurrence}
\end{align}
Here, $\beta_{t\gi d}$ is a learning rate and $\delta_{t\gi d} \coloneqq v_{td} - w_{t-1, \gi d}^{\top} k_t$ is a prediction error, just as in the delta rule \citep{schlag2021linear, yang2024parallelizing}. The responsibilities
\begin{align}
    r_{t \gi d} \coloneqq \frac{\pi_{\gi d}(k_t)\,\exp\!\bigl(-\tfrac{1}{2}\,\pe_{t \gi d}^2\bigr)}{\sum_{\gi'} \pi_{\gi' d}(k_t)\,\exp\!\bigl(-\tfrac{1}{2}\,\pe_{t \gi' d}^2\bigr)}, \label{eq:responsibility}
\end{align}
are the posterior probabilities that the $d$-th dimension of $v_t$ was generated by mixture component~$\gi$. Once the state $w_{t \gi d}$ has been updated, the query $q_t$ then retrieves the output $o_{td}$ according to the mixture of linear regressors, as
\begin{align}
    o_{td} = \sum_{\gi=1}^{\GI} \pi_{\gi d}(q_t) \, w_{t \gi d}^\top q_t, \label{eq:retrieval}
\end{align}
where $\pi_{\gi d}(q_t)$ is also a learned projection from the input $x_t$.

In summary, we derive the \method recursion \eqref{eq:recurrence} and retrieval \eqref{eq:retrieval} from the TTR framework by choosing our model as a mixture of linear regressors and our optimizer as online EM.
We build on this probabilistic perspective in the next two sections: Section~\ref{sec:temporal_sla} adds temporal persistence to the prior over mixture assignments, and Section~\ref{sec:gated_sla} derives input-dependent gating from Gaussian priors on the mixture weights.

\subsection{Regression Model Extension: Temporal \pmethod}
\label{sec:temporal_sla}

A natural benefit of \method is that its mixture components can accommodate different "contexts."
In many settings, context persists across consecutive inputs. 
For example, in an essay comparing the musician Harry Styles with the magician Harry Potter, the discussion of each person likely spans a full paragraph, and the value associated with `Harry' should depend on that context.
Mathematically, this persistence suggests that if input $t$ has a high probability of belonging to mixture component $\gi$, then so should input $t+1$.

However, \method as formulated in \Cref{ssc:sla_mod} does not have any persistence in the mixture assignments.
The mixture assignment of each time step is memoryless, starting over with another draw from its prior $\pi$.
To address this limitation, we extend \method with a Markovian prior that introduces dependencies between $z_{td}$ and $z_{t+1,d}$.

The standard probabilistic approach for such temporal persistence is a hidden Markov model (HMM), but such an approach would require marginalization and computation of normalizing constants at every time step.
Instead, we use a computationally tractable approximation to HMM filtering.
We maintain a running probability distribution over mixture components for each output dimension $d$, and at each time step, take a convex combination of this distribution with a new input-dependent distribution generated from the current input.
With this extension, the mixture assignments in \method can persist across consecutive time steps.
See \Cref{app:tsla_details} for full details on this temporal recurrence.
\subsection{Regression Model Extension: Gated \pmethod}
\label{sec:gated_sla}

Input-dependent state decay, or gating, has driven strong empirical gains in linear attention, as in GDN \citep{yang2025gdn} and KDA \citep{team2025kimi}.
We present two formulations of Gated \pmethod, which differ in whether the gate is applied after or before computing the responsibilities.
See \Cref{app:gsla_details} for full details.

Under the TTR framework, state decay is equivalent to weight regularization of the fast weights \citep{wang2025test}.
Accordingly, we place Gaussian priors on \pmethod's mixture weights, $w_{jd} \sim \mathcal{N}(0, \lambda_{tjd}^{-1} I)$, which introduces a regularization term to eq.~\eqref{eq:loglik}:
\begin{align}
    \ell_{td} = \log \sum_{\gi=1}^{\GI} \pi_{\gi d}(k_t)\,\exp\!\left(-\tfrac{1}{2} \bigl(v_{td} - w_{\gi d}^\top k_t\bigr)^2\right) - \sum_{\gi=1}^{\GI} \tfrac{\lambda_{t\gi d}}{2} \lVert w_{\gi d} \rVert_2^2 + c. \label{eq:gated_loglik}
\end{align}
A one-step stochastic gradient ascent on eq.~\eqref{eq:gated_loglik} yields the recurrence
\begin{equation}
w_{tjd} = \alpha_{tjd}\, w_{t-1,jd} + \beta_{tjd}\, r_{tjd}\, \delta_{tjd}\, k_t,
\qquad
\alpha_{tjd} \coloneqq 1 - \beta_{tjd} \lambda_{tjd},
\label{eq:post-decay}
\end{equation}
where we parameterize $\alpha_{tjd} \in [0, 1]$ as a learned, input-dependent gate that is applied after the responsibilities are computed.

Alternatively, following the implementation of the recurrent form of GDN \citep{yang2025gdn, yang2024fla}, which applies the gate before computing the prediction error, we obtain the recurrence
\begin{equation}
w_{tjd} = \alpha_{tjd}\, w_{t-1,jd} + \beta_{tjd}\, r^{-}_{tjd}\, \delta^{-}_{tjd}\, k_t,
\label{eq:pre-decay}
\end{equation}
where $\delta^{-}_{tjd} \coloneqq v_{td} - \alpha_{tjd}\, w_{t-1,jd}^{\top} k_t$ are the prediction errors at the decayed state and $r^{-}_{tjd}$ are the responsibilities computed from these errors.
The two formulations coincide for $\GI = 1$ up to a reparameterization, but differ for $\GI > 1$.
In practice, we use eq.~\eqref{eq:pre-decay} for Gated \pmethod.

\subsection{Improving the Optimization Algorithm: Load Balancing}\label{ssc:load}

Without regularization, \method is susceptible to expert collapse, where a few mixture components capture all responsibility mass, and the remaining experts go unused. 
We employ two complementary strategies to prevent this.

First, we add an auxiliary loss that adapts load balancing loss \citep{shazeer2017outrageously, fedus2022switch}.
We maximize the mutual information $I(\text{token};\,\text{expert}) = H_{\mathrm{marg}} {-} H_{\mathrm{cond}}$ between tokens and expert assignments, where $H_{\mathrm{marg}}$ is the entropy of the time-averaged responsibilities and $H_{\mathrm{cond}}$ is the mean per-token entropy. Maximizing $H_{\mathrm{marg}}$ encourages balanced utilization across experts, while minimizing $H_{\mathrm{cond}}$ encourages per-token specialization. The auxiliary loss is,
\begin{equation}
  \mathcal{L}_{\mathrm{MI}} = - \lambda_{\mathrm{marg}} H_{\mathrm{marg}} + \lambda_{\mathrm{cond}}\, H_{\mathrm{cond}}.
\end{equation}
It is computed for both the key-side and query-side responsibilities and averaged.
 
Second, following \citet{shazeer2017outrageously}, we inject learned input-dependent noise into the expert logits during training:
\begin{equation}
  \tilde{\pi}_{\gi d}(x_t) = \frac{\exp\!\bigl(\SW_{\gi d}^\top x_t + \epsilon_{t \gi d} \cdot \mathrm{softplus}(\theta_{\mathrm{noise},\gi d}^\top\, x_t)\bigr)}{\sum_{\gi'} \exp\!\bigl(\SW_{\gi' d}^\top x_t + \epsilon_{t \gi' d} \cdot \mathrm{softplus}(\theta_{\mathrm{noise},\gi' d}^\top\, x_t)\bigr)}, \qquad \epsilon_{t \gi d} \sim \mathcal{N}(0,1).
\end{equation}
Noise is injected into both the key-side and query-side expert logits, as well as into the gate logits in Temporal \method (Section~\ref{sec:temporal_sla}), each with its own learned projection.
\section{Related Work}\label{sec:related_work}

We discuss two closely related linear attention sequence layers to \pmethod: DeltaNet \citep{schlag2021linear, yang2024parallelizing} and Mixture-of-Memories (MoM) \citep{du2026mom} / Sparse State Expansion (SSE) \citep{pan2025sse}.
The similarity of these sequence layers to \method motivates our use of them as our linear attention baselines in \Cref{sec:results}.
We include a more expansive discussion of related works in \Cref{app:x_related_work}.

\subsection{DeltaNet}

DeltaNet \citep{schlag2021linear} is a sequence modeling architecture that uses the \emph{delta rule} for its linear attention sequence mixer: its state $W_t \in \mathbb{R}^{D \times D}$ updates according to \cref{eq:delta_net_recurrence}, and then $W_t$ provides a linear mapping from query to output, as shown in \cref{eq:delta_net_retrieval}:
\begin{align}
    W_t & = W_{t-1} + \beta_t \left(v_{t} - W_{t-1} k_{t} \right) k_{t}^{\top} \label{eq:delta_net_recurrence} \\
    o_t & = W_t q_t \label{eq:delta_net_retrieval} 
\end{align}

On the one hand, there are clear differences between \method and DeltaNet. The \method recurrence \eqref{eq:recurrence} is nonlinear while the delta rule \eqref{eq:delta_net_recurrence} is linear in its fast-weights $W_{t-1}$. Furthermore, for retrieval, DeltaNet provides a single linear mapping from query to output \eqref{eq:delta_net_retrieval}, while \method switches adaptively between mixtures of linear regressors \eqref{eq:retrieval}. In TTR, DeltaNet and \method correspond to different function classes for the regressor.

On the other hand, in TTR, DeltaNet and \method both use SGD as the optimizer.
The delta rule optimizes $\| v_t - W_t k_t\|^2$, while \method maximizes the marginal log-likelihood \eqref{eq:loglik}. 
This similarity is why the prediction error appears in both recurrences. 
Moreover, it means that DeltaNet's chunkwise parallel algorithm \citep{yang2024parallelizing} could be applied within each Newton iteration to parallelize \method across the sequence length (\Cref{app:chunk}).

\paragraph{Chunkwise parallel algorithm}

\pmethod's nonlinear recurrence in its fast weights $W_t$ can theoretically be parallelized using \emph{parallel Newton iterations}~\citep{lim2024parallelizing, gonzalez2024parallelizing, gonzalez_thesis, danieli2025pararnn}. 
Parallel Newton iterations work by iteratively linearizing the recurrence, and then using parallel compute to evaluate the linearized system, and they are guaranteed to converge \citep{gonzalez_thesis}.
Previous implementations of these iterations relied on parallel associative scans \citep{blelloch1990prefix}, which incur a prohibitive $O(T D^2)$ memory cost by materializing all hidden states \citep{yang2024deltanetblog2}. To overcome this, we observe that each Newton iteration applied to \method shares the same rank-one update structure as the delta rule recurrence, enabling a novel \emph{chunkwise parallel Newton algorithm} that avoids the parallel scan's memory bottleneck. Although our current experiments rely on a fast recurrent (sequential) Triton kernel for \pmethod, the systems-level optimization of this chunkwise parallel approach presents an interesting direction for future work. We discuss parallelization further in \Cref{app:chunk}.

\subsection{Mixture-of-Memories}

The most similar method to \method is the Mixture-of-Memories (MoM) layer introduced by \citet{du2026mom}.
MoM also splits the hidden state into $\GI$ different memory states, and uses a learnable router to select which memories to activate based on the input.
The Sparse State Expansion (SSE) layer of \citet{pan2025sse} also uses a mixture of memories, albeit with sparse updates.
Since the routers of these layers do not depend on the previous state, their recurrences are \emph{linear} in their hidden states $W_t$.
In contrast, we derive our \method update from the TTR framework.
Consequently, our ``router'' is based on the posterior responsibilities, which are functions not only of the input but also the prior state, rendering our recurrence \emph{nonlinear} in $W_t$. 
Increasingly, theoretical \citep{siems2026learning, merrill2026linear} and empirical \citep{farsang2025scaling, zattra2025context, mishra2026m} 
results demonstrate that nonlinear recurrences have superior expressivity to linear recurrences.

\section{Experimental Results}
\label{sec:results}

We evaluate \method in three settings.
Section~\ref{sec:regression} highlights \pmethod's ability to approximate the input--output mapping of a single \softmax attention head.
Section~\ref{sec:synthetic} evaluates \method on two synthetic benchmarks: context-dependent associative recall and in-context language learning. 
Section~\ref{sec:language-modeling} scales to language models pretrained on FineWeb-Edu, with downstream evaluation on commonsense reasoning and recall-intensive tasks.
Complementing these evaluations, \Cref{sec:throughput-comparison} profiles training and inference efficiency, and \Cref{sec:ablation-study} ablates \pmethod's key design choices.

For our language modeling evaluation, we compare \method against a suite of ungated and gated baselines: DeltaNet \citep{schlag2021linear, yang2024parallelizing}, Gated DeltaNet (GDN) \citep{yang2025gdn}, KDA \citep{team2025kimi}, and MoM \citep{du2026mom} with DeltaNet-style (DN-MoM) and GDN-style (GDN-MoM) updates, as well as Transformer++ \citep{touvron2023llama} and FoX \citep{lin2025forgetting} for softmax attention variants.
We further compare Hybrid GDN and Hybrid Temporal \pmethod, which interleave three recurrent layers with one softmax attention layer following \citet{team2025kimi}.
The recurrent state size is matched across all recurrent models in every experiment.
See \Cref{appendix:exp} for full experimental details.

\subsection{Approximating Softmax Attention}
\label{sec:regression}

We test whether \method can better approximate the input–output mapping of a single \softmax attention head than DeltaNet, given the same recurrent state size.

\paragraph{Setup.}
We construct a synthetic regression task in which the target function is causal \softmax attention.
In this experiment, we set $D=32$.
Inputs $x_t \in \mathbb{R}^{D}$ are drawn from 32 randomly sampled clusters in $\mathbb{R}^{D}$.
Using fixed, randomly sampled projection matrices $\SW_Q, \SW_K, \SW_V$, we compute queries $q_t = \SW_Q x_t$, keys $k_t = \SW_K x_t$, and values $v_t = \SW_V x_t$.
The regression targets are the outputs of causal \softmax attention with temperature $\tau = \sqrt{D} / 2$:
\begin{equation}
    o_t = \sum_{i=1}^{t} \frac{\exp(\, q_t^\top k_i / \tau )}{\sum_{j=1}^{t} \exp( q_t^\top k_j / \tau )}\, v_i.
\end{equation}
We generate 65{,}536 training sequences of length 512 and report test $R^2$ on a held-out set of 2{,}048 sequences.
We compare DeltaNet and \method at matched state budgets $B = H \times \GI$, where $H$ is the number of heads and $\GI$ the number of mixtures per head ($\GI=1$ recovers DeltaNet).
All configurations use a fixed head dimension of $D = 32$, giving a total recurrent state size of $B \times D^2$.
For different factorizations $(H, \GI)$, we expand the key and value dimensions to keep $D$ fixed, ensuring that both models maintain the same total recurrent state size.
For each budget $B \in \{1, 2, 4, 8, 16, 32\}$, we sweep all factorizations $(H, \GI)$ with $H \times \GI = B$ and $\GI \le 16$ for \pmethod, and set $H = B,\, \GI=1$ for DeltaNet.
For a given state budget, many additional configurations become available if the head dimension $D$ is also allowed to vary. 
For instance, a budget of $H \times \GI \times D^2 = 4 \times 1 \times 32^2 = 4{,}096$ can equivalently be realized by $H=4$ heads with $\GI=4$ mixtures and keys and values projected down to dimension $d=16$, since $4 \times 4 \times 16^2 = 4{,}096$. 
We fix $D = 32$ for this experiment. 

\paragraph{Results.}
Figure~\ref{fig:sla_schematic}C plots test $R^2$ against recurrent state size.
At every state budget, the best \method configuration achieves a higher $R^2$ than DeltaNet.
At $B{=}2$, the one-head \method (\texttt{H1\GI2}) outperforms the two-head DeltaNet (\texttt{H2}) by a clear margin, and this advantage persists as the budget increases.
This indicates that allocating budget to mixtures within a head, rather than to independent heads, better approximates \softmax attention.

As the budget grows, the best \method configurations tend to balance heads and mixtures (e.g., \texttt{H2\GI2} at $B{=}4$) rather than concentrating all budget in either dimension. Both forms of capacity---parallel heads and per-head mixtures---play complementary roles.

In~\Cref{fig:regression_all}, we plot all \method configurations at each budget, where the right panel shows the number of model parameters on the x-axis.
Overall, these results confirm that \method provides a more expressive function class for approximating \softmax attention than the independent-head decomposition used by DeltaNet.

\subsection{Synthetic Benchmarks}
\label{sec:synthetic}

\begin{figure}[!t]
    \centering
    \begin{subfigure}[t]{0.49\textwidth}
        \centering
        \includegraphics[width=\textwidth]{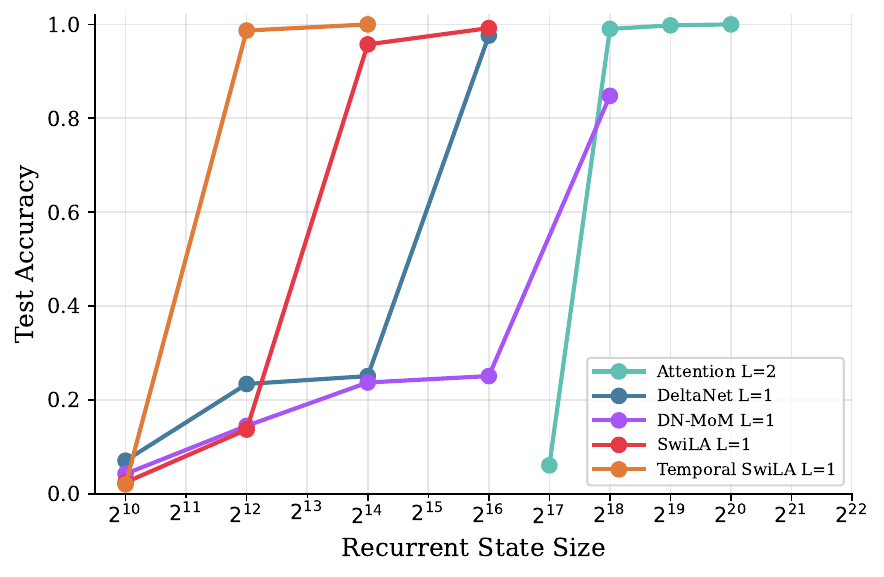}
        \caption{Context-Dependent Associative Recall}
        \label{fig:cdar}
    \end{subfigure}
    \hfill
    \begin{subfigure}[t]{0.49\textwidth}
        \centering
        \includegraphics[width=\textwidth]{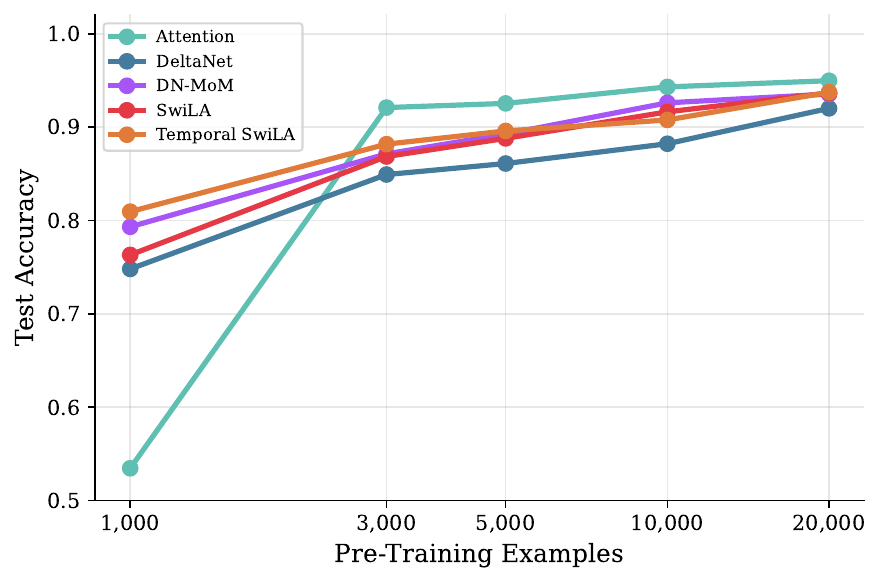}
        \caption{RegBench}
        \label{fig:regbench}
    \end{subfigure}
    \caption{\textbf{Synthetic Benchmarks}.
    (\subref{fig:cdar})~Context-Dependent Associative Recall with 4 contexts and 32 key-value pairs per context: test accuracy vs.\ state size.
    (\subref{fig:regbench})~RegBench: test accuracy as a function of training set size.}
    \label{fig:synthetic}
\end{figure}

We evaluate \method on two synthetic benchmarks: context-dependent associative recall and in-context language learning benchmarks. 

\paragraph{Context-Dependent Associative Recall (CDAR).}
Adapted from the associative recall task of \citet{zoology2024}, we design a variant that tests whether a model can maintain multiple separate associative memories and persist context information to select the correct one at retrieval time.
In CDAR, the same key maps to different values depending on a context token.
Recurrent models must therefore contend with interference across contexts and retain which context is active when answering queries.
Crucially, since the context token appears only once at the start of each block and the query region is separated from the context blocks, the task requires temporal persistence of context awareness that cannot be resolved by the short convolutions commonly prepended to recurrent layers.
Each example contains $C$ context blocks followed by a query region.
A context block begins with a context token $c_i$ and lists key-value pairs under that context.
In the query region, each context's queries are grouped contiguously: the context token appears once, followed by keys whose values the model must predict.
For instance, with $C = 2$ contexts and 3 key-value pairs each:
\begin{align*}
    \underbrace{\mathbf{c_1}, k_2, v_{1,2}, k_1, v_{1,1}, k_3, v_{1,3}
    \;|\; \mathbf{c_2}, k_3, v_{2,3}, k_2, v_{2,2}, k_1, v_{2,1}}_{\text{context blocks}}
    \;|\;
    \underbrace{\mathbf{c_2}, k_3, \mathbf{?}, k_1, \mathbf{?}, k_2, \mathbf{?}
    \;|\; \mathbf{c_1}, k_2, \mathbf{?}, k_3, \mathbf{?}, k_1, \mathbf{?}}_{\text{queries}}
\end{align*}
The key set is shared across all contexts, but each (context, key) pair maps to a distinct value.
The model must use the context token to determine which mapping is active and carry that information forward to the answer positions. Loss is computed only at answer positions.
We use $C=4$ contexts with 32 key-value pairs each (128 total associations), vocabulary size 8{,}192, and sequence length 1{,}024.
Figure~\ref{fig:cdar} shows the best test accuracy across learning rates and random seeds.
Temporal \method with a single head with dimension $D=32$ and 4 mixtures (recurrent state size of $2^{12}$) is the smallest configuration that achieves near-perfect accuracy, suggesting that the mixtures with temporal priors are crucial.
All other recurrent models (nearly) solve the task with one layer as the state size increases, while the Transformer requires two, consistent with recent findings that single-layer Transformers cannot perform associative recall beyond chance \citep{okpekpe2025revisiting}.

\paragraph{RegBench: In-Context Language Learning.}
RegBench~\citep{akyurek2024context} tests how efficiently different architectures learn to perform in-context language learning using probabilistic finite automata (PFAs), where they found that Transformers outperformed recurrent models.
Each input packs 10–20 example strings from a randomly generated PFA (4-12 states, alphabet size 4-18) into a single sequence, and the model must learn the PFA's next-token distribution in-context via next-token prediction.
We compare five 8-layer models: \method and Temporal \pmethod, DN-MoM, DeltaNet, and a Transformer.
We report accuracy on a held-out test set of 500 PFAs using the best checkpoint by validation loss. 
As shown in Figure~\ref{fig:regbench}, Temporal \method and DN-MoM exhibit strong sample efficiency, outperforming the Transformer in the low-data regime (1,000 training PFAs), while the Transformer's accuracy improves steeply with more data and ultimately reaches the highest accuracy.
All models achieve similar performance at scale, with DeltaNet slightly trailing.
\subsection{Pretraining Language Models}\label{sec:language-modeling}

We evaluate \method on a realistic language modeling pipeline.
All models have 374M parameters, share the same architecture skeleton (24 layers and 1024 hidden size), and are trained on 15B tokens from FineWeb-Edu \citep{penedo2024fineweb}.
\method and its variants show strong results across all tasks, narrowing the gap to Transformer++ and even surpassing it on several benchmarks.

\begin{table}[t!]
\centering
\caption{\textbf{Commonsense Reasoning}. Best result per task in \textbf{bold}, second best \underline{underlined}.}
\label{tab:commonsense}
\resizebox{\textwidth}{!}{%
\begin{tabular}{@{}l @{\hspace{1em}} cc @{\hspace{1em}} ccccccccc@{}}
\toprule
& \multicolumn{2}{c}{\textbf{Perplexity} ($\downarrow$)}
& \multicolumn{9}{c}{\textbf{Accuracy} ($\uparrow$, \%)} \\
\cmidrule(lr){2-3} \cmidrule(lr){4-12}
\textbf{Model (374M)}
  & Wiki.
  & LMB.
  & LMB.
  & PIQA
  & Hella.
  & Wino.
  & ARC-E
  & ARC-C
  & SIQA
  & BoolQ
  & Avg. \\
\midrule
Transformer++ & \textbf{28.2} & \textbf{37.4} & \textbf{33.5} & \textbf{66.0} & \textbf{32.4} & \textbf{52.6} & 56.0 & \textbf{24.1} & \textbf{38.4} & \textbf{59.1}& \textbf{45.3} \\
FoX      & 28.7 & 50.3 & 30.8 & 65.2 & 32.2 & 52.5 & \textbf{57.2} & 23.2 & 37.9 & 58.0 & 44.6 \\
\midrule
DeltaNet      & 29.8 & 46.3 & 27.9 & 63.9 & 31.9 & \underline{51.6} & 55.6 & 21.8 & 38.3 & 58.3 & 43.7 \\
Gated DeltaNet      & \underline{27.7} & {35.7} & {32.0} & 66.0 & \underline{33.3} & \textbf{51.8} & \underline{57.5} & {24.6} & \textbf{39.0} & 55.8 & 45.0 \\
KDA           & \textbf{23.4} & \textbf{26.8} & \textbf{36.4} & \textbf{67.0} & \textbf{34.9} & 51.5 & \textbf{60.7} & \textbf{27.3} & \underline{38.9} & 56.2 & \textbf{46.6}\\
DN-MoM           & 40.0 & 84.5 & 24.1 & 62.6 & 30.4 & 49.2 & 52.2 & 22.8 & 36.6 & \textbf{62.0} & 42.5 \\
GDN-MoM           & 35.3 & 56.9 & 27.5 & 64.6 & 31.6 & 50.2 & 54.2 & 23.6 & 37.2 & 53.3 & 42.8 \\
\method  & 28.8 & 43.0 & 30.4 & \underline{66.6} & 32.0 & 51.5 & 56.1 & 22.8 & 38.1 & 60.1 & 44.7 \\
Gated \method  & 27.8 & 31.8 & \underline{34.6} & 66.1 & 32.7 & 50.4 & \underline{57.5} & \underline{25.0} & 38.1 & 59.1 & \underline{45.4} \\
Temporal \method           & 28.6 & 39.4 & 31.5 & 66.5 & 31.7 & 51.1 & 57.2 & 23.2 & 38.8 & \underline{61.5} & {45.2} \\
Gated Temporal \method  & 27.8 & \underline{31.7} & 34.5 & 65.8 & 32.9 & \textbf{51.8} & 56.5 & 24.5 & 38.5 & 58.4 & \underline{45.4} \\
\midrule
Hybrid GDN & 31.6 & 56.3 & 28.5 & 64.8 & 32.2 & \textbf{50.7} & 54.6 & 23.1 & \textbf{39.0} & \textbf{61.4} & 44.3 \\
Hybrid Temporal \method  & \textbf{27.8} & \textbf{40.0} & \textbf{32.4} & \textbf{66.1} & \textbf{32.6} & 50.2 & \textbf{57.2} & \textbf{23.3} & 37.7 & 61.1 & \textbf{45.1} \\
\bottomrule
\end{tabular}%
}
\end{table}

\paragraph{Commonsense Reasoning.}
Following \cite{gu2024mamba}, we evaluate our model on a suite of commonsense reasoning tasks: Wikitext \citep{merity2016pointer}, LAMBADA \citep{paperno2016lambada}, PIQA \citep{Bisk2020}, HellaSwag \citep{zellers2019hellaswag}, WinoGrande \citep{sakaguchi2019winogrande}, ARC-easy and ARC-challenge \citep{Clark2018ThinkYH}, SIQA \citep{sap2019social}, and BoolQ \citep{clark2019boolq}.
Among the linear attention layers, KDA achieves the strongest commonsense results. 
Every \method variant attains higher average accuracy than DeltaNet, DN-MoM and GDN-MoM, and the Gated \method variants additionally exceed GDN, slightly surpassing Transformer++ on average and outperforming it on several individual benchmarks.

\paragraph{Recall-Intensive Tasks.}

To measure the ability to perform context-based retrieval and comprehension, we evaluate on six recall-intensive tasks following \cite{arora2024just}: SWDE \citep{lockard2019openceres, arora2023language}, FDA \citep{arora2023language}, SQuAD \citep{rajpurkar2018know}, TriviaQA \citep{joshi2017triviaqa}, NQ \citep{kwiatkowski2019natural}, and DROP \citep{dua2019drop}.
Table~\ref{tab:recall} shows that the softmax attention models achieve the highest scores, with FoX attaining the best average, benefiting from their full KV caches that grow linearly with context length. 
Every \method variant outperforms DeltaNet and both MoM variants on all six recall benchmarks, and Gated \method variants achieve the highest averages among all recurrent models, including KDA. 
Among the hybrid models, Hybrid Temporal \method surpasses both Hybrid GDN and Transformer++, approaching the recall performance of FoX. 
These gains confirm that \pmethod's mixture structure meaningfully improves associative recall compared to a single linear regressor, narrowing the gap to softmax attention while retaining a fixed-size recurrent state.

\subsection{Throughput Comparison}\label{sec:throughput-comparison}

We profile the training and inference throughput and peak memory of six 1.3B-parameter models, a scale widely adopted for such analyses \citep{gu2023mamba, yang2025gdn}: Transformer, GDN, Hybrid GDN, \pmethod, Temporal \pmethod, and Hybrid Temporal \pmethod.
For the hybrid models, we follow the 3:1 layer ratio of \citet{team2025kimi}, interleaving three recurrent layers with one Transformer layer.
See \Cref{app:throughput} for details.

\begin{table}[t!]
\centering
\caption{\textbf{Recall-Intensive Tasks}. All scores use the \emph{contains} metric (\%, $\uparrow$). Best in \textbf{bold}, second best \underline{underlined}. All inputs are truncated to 2K tokens.}
\label{tab:recall}
\footnotesize
\setlength{\tabcolsep}{3pt}
\begin{tabular}{@{}lccccccc@{}}
\toprule
\textbf{Model} & SWDE & FDA & SQuAD & TriviaQA & NQ & DROP & Avg. \\
\midrule
Transformer++ & {28.4} & {29.5} & \textbf{31.5} & {46.0} & \textbf{15.6} & {18.9} & {28.3} \\
FoX      &     \textbf{36.4}       &     \textbf{53.0}      &     30.6     &   \textbf{46.4}        &     \textbf{15.6}    &  \textbf{20.2}   &   \textbf{33.7}    \\
\midrule
DeltaNet      & 7.9           & 5.4          & 23.4          & 41.7          & 10.5          & 14.9    & 17.3      \\
Gated DeltaNet      & 9.1           & 5.5          & 25.2          & 44.4          & {12.9}          & \underline{18.0}    & 19.2      \\
KDA      & \textbf{13.9}           &    5.5       &    {25.5}      &   \textbf{46.8}       &    12.5       &  {17.6}   &  {20.3}   \\
DN-MoM           & 5.4          & 2.1          & 21.2          & 37.0          & 9.1          & 14.8 &   14.9          \\
GDN-MoM           & 8.3          & 3.8          & 18.4          & 41.7          & 10.0          & 16.3 &   16.4          \\
\method           & 12.9          & 8.0          & 24.7          & 42.5          & {12.9}     & {17.6}   & {19.7}      \\
Gated \method           & 10.2          & \underline{10.1}          & \textbf{26.3}          & \underline{45.2}          & \underline{13.6}     & 17.5    & \underline{20.5}      \\
Temporal \method           & \underline{13.6}          & \textbf{11.6}          & 23.8          & 42.8          & {13.3}          & 17.2    & {20.3}      \\
Gated Temporal \method           &   12.0       &      7.5     &    \underline{26.0}      &   44.9       &    \textbf{13.8}       &   \textbf{19.2}  &  \textbf{20.6}   \\
\midrule
Hybrid GDN           & 25.0          & 38.2          & \textbf{34.9}          & 43.7          & \textbf{16.0}     & \textbf{19.1}    & 29.5      \\
Hybrid Temporal \method           & \textbf{32.3}          & \textbf{49.1}          & 30.0          & \textbf{45.8}          & 14.9     & 18.9    & \textbf{31.8}      \\
\bottomrule
\end{tabular}
\end{table}

Overall, \method achieves 4--5$\times$ higher end-to-end inference throughput than the Transformer baseline, as it does not require a KV cache and can therefore accommodate substantially larger batch sizes.
Relative to GDN, \method is approximately 2$\times$ slower at training time and 1.25--1.5$\times$ slower at inference time, and it has a larger memory footprint.
We expect that part, though not all, of this gap can be narrowed through further engineering, such as custom CUDA kernels or the parallelization of the recurrence discussed in \Cref{app:chunk}.
\subsection{Ablation Study}\label{sec:ablation-study}

We conduct ablation experiments that isolate the contribution of the number of experts, temporal persistence, and the load-balancing loss.
We train 115M-parameter \method variants on 5B tokens and report test perplexity on 100M held-out tokens in Figure~\ref{fig:router_diagnostics}, alongside routing diagnostics defined in \Cref{app:routing}.
We highlight three findings.
First, temporal persistence consistently improves perplexity over vanilla \pmethod.
Second, load balancing loss is critical for vanilla \pmethod, which otherwise suffers severe expert collapse, but has a negligible effect on Temporal \pmethod; temporal persistence itself appears to act as a natural regularizer against collapse.
Third, under a fixed recurrent state size budget, more experts do not necessarily help: increasing the number of experts from 4 to 16 slightly degrades performance, since each expert must then use smaller query, key, and value projections, creating a tradeoff between expert count and per-expert projection capacity.
\section{Conclusion}

We introduce \textit{Switching Linear Attention} (\pmethod), a sequence layer that retains a recurrent state whose size is independent of sequence length as in linear attention, while representing a nonlinear map from queries to outputs as in softmax attention.
Building on the test-time regression (TTR) framework \citep{wang2025test}, which unifies the design of sequence layers around associative recall, we derive \method from a mixture of linear regressions via online expectation--maximization.
Across associative recall, in-context language learning, and language modeling benchmarks, \method performs strongly and narrows the gap to softmax attention, even surpassing it in several settings.

\paragraph{Limitations}
While \method increases the expressivity of standard linear attention, it retains a fixed state size. As a result, its capacity cannot grow with the context length, unlike the KV cache of softmax attention\footnote{Though even softmax attention struggles with attention rot in long contexts \citep{hsieh2024ruler}.}.
Moreover, as \method uses a nonlinear recurrence, we currently pretrain \method with sequential evaluation across the sequence length, whereas both softmax and linear attention mechanisms are designed to allow for training that is parallelized across the sequence length.
While we can parallelize the nonlinear recurrence of \method using multiple Newton iterations, further optimizations are required to design custom kernels that yield speed-ups over optimized kernels for sequential evaluation.

\paragraph{Future Outlook}

\method demonstrates the promise of TTR: we can design sequence layers with desired in-context learning properties by choosing appropriate regression models and update algorithms.
We anticipate TTR to open up a range of possibilities between the efficiency of linear attention and the expressivity of softmax attention. 
For example, while \method shows the promise of mixture modeling as occupying a useful middle ground between these two extremes, other function classes such as local linear regression \citep{zuo2026local} may prove useful in fully exploring this space.
Moreover, \pmethod's enhanced ability to approximate softmax attention (Figure 1C) while retaining a fixed-state size appears promising for distillation---converting pretrained softmax layers into recurrent layers for efficient inference \citep{bick2024transformers, wang2024mamba, zhang2024lolcats}.

\newpage

\section*{Acknowledgments}
H.D.L. is supported by the Ketterer-Vorwald Stanford Interdisciplinary Graduate Fellowship.
N.Z. is supported by Postdoc.Mobility grant P500-2\_235376 from the Swiss National Science Foundation.
E.B.F. is supported in part by ONR Grant N00014-22-1-2110 and the Stanford Institute for Human-Centered Artificial Intelligence (HAI), and is a Biohub, San Francisco, Investigator.
S.W.L. is supported by grants from the NIH (U01NS136507, R01NS131987, R01NS113119, RF1MH133778, R01AG097491, R01NS130789), the NSF (2440859), and the Simons and McKnight Foundations.
We thank the members of the Linderman Lab and the Dynamode Lab for feedback throughout this project.
We are also grateful for GPU support from the Stanford Research Computing Center's Sherlock cluster, from VESSL AI, and from Marlowe \citep{marlowe2025}, Stanford University's GPU-based Computational Instrument, supported by Stanford HAI and Stanford Research Computing.

\newpage

\bibliography{refs}
\bibliographystyle{colm2026_conference}

\newpage
\appendix
\crefalias{section}{appendix}
\crefalias{subsection}{appendix}
\renewcommand{\thefigure}{S\arabic{figure}}
\setcounter{figure}{0}

\section{Extended Background}
\subsection{Design Space of Test-Time Regression}
\label{app:ttr_design_space}

The test-time regression framework of~\citet{wang2025test} recovers a broad class of existing sequence layers through different design choices. 
Linear attention~\citep{katharopoulos2020transformers} corresponds to linear regression with a whitened design matrix approximation. 
Adding decaying regression weights yields gated linear attention and state-space models~\citep{yang2023gated, dao2024transformers}. 
Learning the regression in an online streaming manner with stochastic gradient descent produces DeltaNet~\citep{schlag2021linear, yang2024parallelizing}, with further variations recovering Longhorn~\citep{liu2024longhorn}, Gated
DeltaNet~\citep{yang2025gdn}, and Titans~\citep{behrouz2024titans}. 
Softmax attention arises as a nonparametric local constant regressor.

\subsection{Mixtures of Linear Regressions}
\label{app:mixture_background}

Consider a dataset $\{(x_t, y_t)\}_{t=1}^T$ with $x_t \in \mathbb{R}^D$ and $y_t \in \mathbb{R}^D$, assumed to be generated from a mixture of $\GI$ linear Gaussian regression models. 
Let $z_t \in \{1, \ldots, \GI\}$ denote a latent component indicator with mixing weights $\pi = (\pi_1, \ldots, \pi_\GI)$, where $\pi_\gi > 0$ and $\sum_{\gi=1}^\GI \pi_\gi = 1$.

Conditioned on $z_t = \gi$, the output $y_t$ is generated according to
\begin{equation}
  y_t \mid x_t, z_t = \gi \sim \mathcal{N}(W_\gi x_t, \Sigma_\gi),
\end{equation}
where $W_\gi \in \mathbb{R}^{D \times D}$ is the regression matrix and $\Sigma_\gi \in \mathbb{R}^{D \times D}$ is the noise covariance, often restricted to the isotropic form $\Sigma_\gi = \sigma_\gi^2 I_D$. 
The complete parameter set is $\Theta = \{\pi_{\gi}, W_{\gi}, \Sigma_{\gi} \}_{\gi=1}^{\GI}$.

Given parameters $\Theta$, inference over the latent variables is performed by computing the responsibilities
\begin{equation}
  r_{tm} = \frac{\pi_m \,\mathcal{N}(y_t;\, W_m x_t,\, \Sigma_m)}
    {\sum_{\gi^\prime=1}^\GI \pi_{\gi^\prime} \,\mathcal{N}(y_t;\, W_{\gi^\prime} x_t,\, \Sigma_{\gi^\prime})},
\end{equation}
which represent the posterior probability that observation $(x_t, y_t)$ was generated by component $\gi$. 
The model parameters and responsibilities can be estimated using the Expectation--Maximization (EM) algorithm, which alternates between computing the responsibilities (E-step) and maximizing the expected complete-data log-likelihood with respect to $\Theta$ (M-step).

\clearpage
\section{Model Specification Details}\label{app:tech_details}

\subsection{Derivation of \method Recurrence}\label{app:sla_recurrence_derivation}

In this appendix, we work out in more detail the derivation of the \method recurrence in \cref{eq:recurrence}.

To make bookkeeping easier, we define
\begin{align} 
    \pi_{\gi} & := \pi_{\gi d}(k_t)\,\exp\!\bigl(-\tfrac{1}{2}\,\pe_{t \gi d}^2\bigr) \label{eq:pi_m}  \\
    Z & := \sum_{\gi} \pi_{\gi} \label{eq:Z}
\end{align}

Taking derivatives, it follows that
\begin{align*}
    \dfrac{\partial \pi_\gi}{\partial w_{t-1,\gi d}} & = \pi_{\gi d}(k_t) \exp(- \pe_{t \gi d}^2 / 2) \pe_{t \gi d} k_t \\
    & = \pi_{\gi} \pe_{t \gi d} k_t \\
    \dfrac{\partial Z}{\partial w_{t-1, \gi d}} & = \dfrac{\partial \pi_{\gi}}{\partial w_{t-1,\gi d}} \\
\end{align*}

Thus, we obtain
\begin{align*}
    \dfrac{\partial \ell_{td}}{\partial w_{t-1, \gi d}} & = \frac{\nicefrac{\partial \pi_{\gi} }{\partial w_{t-1, \gi d}}}{Z} \\
    & = r_{t \gi d} \pe_{t \gi d} k_t. \quad \quad \blacksquare
\end{align*}

\subsection{Details of the \method Layer}\label{app:sla_details}

In the practical implementation of \pmethod, we obtain many quantities of interest as learnable projections of the input $x_t \in \mathbb{R}^{d_{\text{in}}}$. 

In keeping with standard practice, we generate keys, queries, and values as projections of the input $x_t$, according to
\begin{align*}
    q_t & = \theta_Q x_t \\
    k_t & = \theta_K x_t \\
    v_t & = \theta_V x_t.
\end{align*}
These projections can also have nonlinearities applied to them, such as the sigmoid linear unit $\mathrm{SiLU}(x) \coloneqq x \sigma(x)$. For language model pretraining (\Cref{sec:language-modeling}), we apply $\mathrm{SiLU}$ to the queries, keys, and values.

Up to reshaping, the learning rate $\beta$ comes from a linear mapping from the inputs, followed by a sigmoid function to keep the learning rate in $[0,1]$.
Similarly, our priors on the keys $\pi(k_t)$ and queries $\pi(q_t)$ for the mixture memberships come from linear projections applied to the inputs $x_t$, followed by the application of softmax over the mixture-component dimension (to make sure the priors are valid probabilities).
In language model pretraining, we use a low-rank projection to reduce the number of parameters used by these projections.

\subsection{Temporal \method Details}\label{app:tsla_details}

We provide more details on how we encourage temporal persistence in our priors over mixture component assignment $\pi_d(k_t)$ and $\pi_d(q_t)$.

In our implementation, the temporal recurrence for both these priors is a convex combination of the previous filtered posterior and a new input-dependent distribution. The relative weighting is controlled by $g$, which is a learned function of the input $x_t$, dictating the amount of persistence of stickiness in our distribution over mixture component assignment. The range of $g$ is $[0,1]$: if $g$ is close to 1, then we use a new probability distribution suggested by the input, while if $g$ is close to 0, at time $t$ we use a very similar probability distribution to that used at time $t-1$.

To make this recursion formal, for time $t$ we define $\bar{r}_{td}$ to be our probability distribution over mixture components for the keys, and $\bar{\rho}_{td}$ to be our probability distribution over the mixture components for the queries. Using these definitions, we can then define the recurrences as
\begin{align}
  g_{td}^{k} &= \sigma \bigl((\theta_g^{k})^\top x_t + b_g^k \bigr), \quad   g_{td}^{q} = \sigma \bigl((\theta_g^{q})^\top x_t + b_g^q \bigr) \\ 
  \bar{r}_{t \gi d} &= (1 - g^{k}_{td})\, r_{t-1,\gi d} + g^{k}_{td}\, \bigl(\pi_{\gi d}^{k}(k_t)\bigr), \\
  \bar{\rho}_{t \gi d}
  &= (1 - g_{td}^{q})\, \bar{\rho}_{t-1,\gi d}
  \;+\; g_{td}^{q}\, \!\bigl(\pi_{\gi d}^{q}(q_t)\bigr),
\end{align}
where $\pi_{ d}^{k}(k_t)$ and $\pi_{d}^{q}(q_t)$ come from the learnable projections followed by softmaxes discussed in \Cref{app:sla_details}. Finally, at time $t$, we use $\bar{r}_{t \gi d}$ as our prior for the key mixture assignment in \cref{eq:responsibility}, and we use $\bar{\rho}_{t \gi d}$ as our prior for the query mixture assignment in \cref{eq:retrieval}.

\subsection{Gated \method Details}\label{app:gsla_details}

We provide more details on the equivalence of the two gating formulations in Section~\ref{sec:gated_sla} when $J = 1$, and on the granularity at which the gate can be applied.

\paragraph{Equivalence of the two gating formulations when $J = 1$.}
Eqs.~\eqref{eq:post-decay} and \eqref{eq:pre-decay} differ only in whether the prediction errors and responsibilities are evaluated at the undecayed or the decayed state. 
When $J = 1$, we obtain $r_{t1d} = r^{-}_{t1d} = 1$, so the responsibilities drop out of both updates. Dropping the
index $j$ and writing each as an affine map of $w_{t-1,d}$,
\begin{align}
  \text{eq.~\eqref{eq:post-decay}:}\quad
  w_{td} &= \big(\alpha_{td} I - \beta_{td}\, k_t k_t^\top\big) w_{t-1,d} + \beta_{td} v_{td} k_t, \\
  \text{eq.~\eqref{eq:pre-decay}:}\quad
  w_{td} &= \alpha_{td}\big(I - \beta_{td}\, k_t k_t^\top\big) w_{t-1,d} + \beta_{td} v_{td} k_t .
\end{align}
With $\alpha_{td} > 0$, re-parameterizing $(\alpha_{td}, \beta_{td}, v_{td}) \mapsto (\alpha_{td},\, \beta_{td}/\alpha_{td},\, \alpha_{td} v_{td})$ makes the two equivalent.

\paragraph{Granularity of the gate.}
Section~\ref{sec:gated_sla} places an isotropic Gaussian prior $w_{jd} \sim \mathcal{N}(0, \lambda_{tjd}^{-1} I)$ on each row of the state. 
More generally, we may place a prior $w_{jdd'} \sim \mathcal{N}(0, \lambda_{tjdd'}^{-1})$, where $d'$ indexes the key dimension, so that $w_{jdd'}$ is the weight connecting key dimension $d'$ to output dimension $d$ within mixture component $j$. 
One step of gradient ascent on the resulting objective yields
\begin{equation}
  w_{tjd} = \alpha_{tjd} \odot w_{t-1,jd} + \beta_{tjd}\, r_{tjd}\, \delta_{tjd}\, k_t,
  \qquad
  \alpha_{tjdd'} \coloneqq 1 - \beta_{tjd}\lambda_{tjdd'},
  \label{eq:gate-elementwise}
\end{equation}
where $\odot$ is the Hadamard product. 
Collecting the fast weights into $W_{tj} \in \mathbb{R}^{D \times D}$ with $(W_{tj})_{dd'} = w_{tjdd'}$, such that rows index output dimensions and columns index key dimensions, the granularity of the decay is controlled by restricting how $\lambda_{tjdd'}$ varies over its indices. 
Tying the precision across the output index, $\lambda_{tjdd'} = \lambda_{tjd'}$, gives a right diagonal scaling $W_{t-1,j}\mathrm{Diag}(a_{tj})$, i.e.\ one gate per key dimension, as in KDA \citep{team2025kimi}. 
Tying it across the key index, $\lambda_{tjdd'} = \lambda_{tjd}$, gives a left diagonal scaling $\mathrm{Diag}(\alpha_{tj}) W_{t-1,j}$, i.e.\ one gate per output dimension, which is eq.~\eqref{eq:post-decay}.
The restriction $\lambda_{tjdd'} = \lambda_{tj}$ recovers the scalar per-head decay of GDN \citep{yang2025gdn}.

\clearpage
\section{Extended Related Work}\label{app:x_related_work}

\paragraph{Test-time regression} \method is motivated by the test-time regression (TTR) framework of \citet{wang2025test}, which allows for the principled design of sequence layers based on ICL desiderata. TTR is related to many other important concepts in sequence modeling, including mesa-optimization, test-time training (TTT), and fast-weights.
Mesa-optimization \citep{von2023uncovering} also shows that effective sequence layers optimize a learning objective during ICL, but often emphasizes the role of pretraining in unlocking such capabilities; whereas TTR focuses on how a chosen recurrent update affects ICL regardless of pretraining. 
TTT also considers the recurrent state to be the parameters of a function, but often considers more general functional relationships (like MLPs) or more general losses, such as reconstruction \citep{sun2024learning} or next-token prediction \citep{tandon2025end}, and is pretrained to encourage metalearning \citep{finn2017model}.
The fast-weights paradigm \citep{vondermalsburg1981correlation, hinton1987using, schmidhuber1992learning, irie2021going, irie2025fast} also distinguishes between pretrained ``slow'' weights and test-time ``fast'' weights, but tends to focus more on the fast weights being generated by hypernetworks \citep{ha2016hypernetworks}.

Other papers have built on TTR with different design goals in mind.
For example, \cite{von2023uncovering, von2025mesanet} construct a linear attention sequence layer they call a \emph{mesa-layer}, which uses correct Newton updates for linear regression (instead of the whitened design matrix approximation used in linear attention \citep{katharopoulos2020transformers}) to fully optimize the regression objective.  
\citet{peng2025gated} introduce Gated KalmaNet Attention (GKA), which draws inspiration from Kalman filtering.
\citet{zuo2026local} introduced Local Linear Attention (LLA), which drew on TTR to interpolate between linear and softmax attention via local linear regression; however, like softmax attention, LLA still requires a memory store that scales as $O(TD)$ memory, whereas \method uses a fixed state size like linear attention.
To our knowledge, \method is the first sequence layer to extend the TTR framework by specifically incorporating mixture modeling; though \citet{wang2026distributed} use a collection of TTR agents which communicate with each other via a fixed weight matrix.

\paragraph{Efficient attention} The main motivation for this paper is that standard softmax attention \citep{vaswani2017attention} is highly expressive, but can be computationally expensive: it suffers from a KV cache that grows with the sequence length. In contrast, linear attention mechanisms \citep{zhang2026survey} compress context to a fixed-dimensional state, and so are computationally cheaper---but this fixed state often results in too much compression of the context, limiting long-range modeling capabilities \citep{zoology2024, von2025mesanet, tandon2025end}. This dilemma has been noted by many in the literature \citep{zhang2024hedgehog, han2024bridging, han2024agent}.
One mechanism to increase the expressivity of recurrent architecture is \emph{state expansion}, namely to make the fixed state-size bigger \citep{gu2025tradeoffs, gelada2025scaling}. One perspective on the benefits of \method is that the mixture of regressors is a form of state expansion. 

Another approach is to relax the constraint of fixed state size, and instead allow the state to grow with the sequence length, albeit slower than linearly (as the KV cache does). For example, \citet{guo2026log} introduce log-linear attention, a sequence layer with state size that grows logarithmically in the sequence length.

Finally, an orthogonal approach to efficient attention is to optimize the software implementations of attention \citep{dao2022flashattention, zadouri2026flashattention} or linear attention \citep{yang2024fla, beck2025fla}, with the goal of making them as performant as possible on GPUs. 
Such approaches also include optimizing the software implementations of sequential evaluations of nonlinear RNNs \citep{poppel2025flashrnn, mishra2026m}.

\clearpage
\section{Chunkwise Parallel Algorithms}\label{app:chunk}

DeltaNet admits a chunkwise parallel algorithm \citep{yang2024parallelizing, yang2024deltanetblog2} that parallelizes its linear recurrence \eqref{eq:delta_net_recurrence} over the sequence length. The \method recurrence \eqref{eq:recurrence} is nonlinear in its hidden state, which makes it harder to parallelize. Nevertheless, \method can still be parallelized, at least theoretically, using \emph{parallel Newton iterations} \citep{lim2024parallelizing,gonzalez2024parallelizing,danieli2025pararnn, gonzalez_thesis}.

However, most previous implementations of parallel Newton iterations use the parallel associative scan \citep{blelloch1990prefix}, colloquially known as the \emph{pscan}. A pscan-based implementation would materialize the full state at every time step, which scales very badly for matrix-valued hidden states \citep{yang2024deltanetblog2}.

One of the novel contributions of this paper is that we show that we can parallelize the \emph{nonlinear} \method recurrence with parallel Newton iterations; but where every Newton iteration is parallelized with the \emph{chunkwise parallel algorithm} used for DeltaNet \citep{yang2024parallelizing, yang2024deltanetblog2}.

Nonetheless, in all our experiments, we use an optimized recurrent Triton kernel for \method (i.e. we evaluate the nonlinear recurrence sequentially).
The purpose of this appendix is to show that a chunkwise parallel implementation is also available in principle.
An optimized implementation of this novel chunkwise parallel algorithm for nonlinear recurrences is an interesting direction for future work.

\subsection{Background on Parallelizing Recurrences}\label{sec:background_parallel}

To develop our novel chunkwise parallel algorithm for nonlinear recurrences, we combine two ideas:
\begin{itemize}
    \item the chunkwise parallel algorithm for DeltaNet \citep{yang2024parallelizing}; and
    \item parallel Newton iterations for nonlinear recurrences \citep{lim2024parallelizing, danieli2025pararnn, gonzalez_thesis}
\end{itemize}

\paragraph{DeltaNet's chunkwise parallel algorithm}

Consider a linear recurrence of the form
\begin{align*}
    W_{t} = A_{t}\, W_{t-1} + b_{t}.
\end{align*}
When the transition matrices $A_t$ are unstructured, parallel evaluation via associative
scans \citep{blelloch1990prefix} requires materializing and multiplying dense matrices, which can be prohibitively costly.
\citet{yang2024parallelizing} observe that when $A_t$ has the structure of a rank-one
perturbation of the identity
\begin{align}\label{eq:id_minus_r1}
    A_t = I - \beta_t\, k_t k_t^\top,
\end{align}
the recurrence admits an efficient "chunkwise" algorithm.
This algorithm gets its name because the sequence is divided into "chunks" of size~$C$.
Within each chunk, the dependency among timesteps is captured by a strictly lower
triangular matrix $\Lambda \in \mathbb{R}^{C \times C}$ with entries
\begin{align}
    \Lambda_{s,t} =
    \begin{cases}
        -\beta_s\, k_s^\top k_t & \text{if } s > t, \\
        0 & \text{otherwise},
    \end{cases}
\end{align}
and intra-chunk states are recovered via forward substitution through this triangular dependency structure.
Boundary states are propagated sequentially across the $N \coloneqq T/C$ chunk boundaries, but this cost is amortized over the chunk size.
The algorithm avoids materializing full $D \times D$ transition matrices, operating instead on $C \times C$ matrices of key inner products, and is well suited to modern GPU hardware.

\paragraph{Parallel Newton iterations for nonlinear recurrences}

Many sequence models involve nonlinear state updates
$w_{t} = f_{t}(w_{t-1})$,
which seem like they must be evaluated sequentially.
Recent work~\citep{song2021accelerating, deeppcr, lim2024parallelizing, gonzalez2025unifying, gonzalez_thesis} shows that such recurrences can be parallelized by recasting sequence evaluation as a fixed-point
problem over the full trajectory.
Starting from an initial guess for the \emph{entire trajectory}, often denoted by $w^{(0)}_{1:T}$, fixed-point methods iteratively converge to the true nonlinear rollout $w_{1:T}$.
A broad class of methods, including Newton, quasi-Newton, and Picard iterations,
can be written in the unified form
\begin{align}\label{eq:unif}
    w^{(i+1)}_{t}
    = f_{t}\!\big(w^{(i)}_{t-1}\big)
    + A_{t}^{(i)}\!\left(w^{(i+1)}_{t-1} - w^{(i)}_{t-1}\right),
\end{align}
where $A_{t}^{(i)}$ approximates the Jacobian $\partial f_t / \partial w_{t-1}(w_{t-1}^{(i)})$.
Because each iteration defines a linear dynamical system, the full sequence can be
evaluated using a parallel scan in $\mathcal{O}(\log T)$ depth \citep{blelloch1990prefix, martin2018parallelizing, smith2023s5}.
Moreover, these fixed-point iterations are guaranteed to converge in at most $T$ iterations \citep{Bellen1989, shih2023parallel, tang2024accelerating, gonzalez2024parallelizing}.

Our core insight in combining these two approaches is that if Jacobian approximation $A_t^{(i)}$ in the parallel Newton iterations in \cref{eq:unif} has the identity-minus-rank-one form of the delta rule \cref{eq:id_minus_r1}, then each fixed-point step can be parallelized with DeltaNet's chunkwise parallel algorithm. 

\subsection{Linearizing the \method Recurrence}\label{ssc:parr_sla}
\label{sec:parallel_sla}

To apply parallel Newton iterations to \pmethod, we must first linearize its recurrence \eqref{eq:recurrence}.
For each mixture component $\gi$ and output dimension $d$, we can write the \method recurrence as
\begin{equation*}
    w_{t \gi d} = f(w_{t-1,\gi d}) \coloneqq w_{t-1,\gi d} + \beta_{t \gi d}\,r_{t \gi d}\,\pe_{t \gi d}\,k_t.
\end{equation*}
To linearize this recurrence, we start by computing the Jacobians $\dfrac{\partial f}{\partial w_{t-1, \gi d}} \in \mathbb{R}^{D \times D}$.
We start this process with some intermediate computations using shorthands defined in \Cref{app:sla_recurrence_derivation}, obtaining
\begin{align*}
    \dfrac{\partial \pe_{t \gi d}}{\partial w_{t-1, \gi d}} & = -k_t^{\top} \\
    \dfrac{\partial r_{t \gi d}}{\partial w_{t-1, \gi d}} & = \frac{ Z \nicefrac{\partial \pi_{\gi}}{\partial w_{t-1,\gi d}} - \pi_{\gi} \nicefrac{\partial Z}{\partial w_{t-1, \gi d}}}{Z^2} \\ 
    & = \frac{(Z - \pi_{\gi}) (\pi_{\gi} \pe_{t \gi d} k_t^{\top})}{Z^2} \\
    & = (1- r_{t \gi d}) r_{t \gi d} \pe_{t \gi d} k_t^{\top}
\end{align*}

Using these derivatives and the product rule, it follows that
\begin{align*}
    \dfrac{\partial r_{t \gi d} \pe_{t \gi d}}{\partial w_{t-1, \gi d}} & = r_{t \gi d} \dfrac{\partial \pe_{t \gi d}}{\partial w_{t-1, \gi d}} + \pe_{t \gi d} \dfrac{\partial r_{t \gi d}}{\partial w_{t-1, \gi d}} \\
    & = - r_{t \gi d} k_t^{\top} + r_{t \gi d} (1 - r_{t \gi d}) \pe_{t \gi d}^2 k_t^{\top} \\
    & = -r_{t \gi d} \left(1 - \pe_{t \gi d}^2 (1 - r_{t \gi d}) \right) k_t^{\top}.
\end{align*}

Thus, it follows that
\begin{align}
    \frac{\partial f}{\partial w_{t-1,\gi d}} = I - g_{t \gi d}\,k_t k_t^\top, \label{eq:jacobian}
\end{align}
where
\begin{align}
    g_{t \gi d} = \beta_{t \gi d}\,r_{t \gi d} \bigl[1 - \pe_{t \gi d}^2(1 - r_{t \gi d})\bigr]. \label{eq:jacobian_coeff}
\end{align}
Thus, the Jacobian of the per-expert, per output dimension \method recurrence is a rank-one perturbation of the identity, which is exactly the structure exploited by the chunkwise algorithm of~\citet{yang2024parallelizing}.

Applying the Newton linearization, the $i$-th iteration solves the linear recurrence
\begin{align}
    w^{(i+1)}_{t \gi d} = \bigl(I - g^{(i)}_{t \gi d}\,k_t k_t^\top\bigr)\, w^{(i+1)}_{t-1,\gi d} + k_t\, p^{(i)}_{t \gi d}, \label{eq:linearized}
\end{align}
where the effective input term is
\begin{align}
    p^{(i)}_{t \gi d} := g^{(i)}_{t \gi d}\,k_t^\top w^{(i)}_{t-1,\gi d} + \beta_{t \gi d}\,r^{(i)}_{t \gi d}\,\pe^{(i)}_{t \gi d}. \label{eq:effective_input}
\end{align}
This effective input term follows because we know from \cref{eq:unif} that bias term $b_t$ takes the form $f_{t}(w_{t-1}^{(i)}) - A_t w_{t-1}^{(i)}$.

In this setting, $A_t = I - g_{t \gi d}^{(i)} k_t k_t^{\top}$, and so the bias term is given by
\begin{align*}
    b_t^{(i)} & = w_{t-1, \gi d}^{(i)} + \beta_{t \gi d} r_{t \gi d}^{(i)} \pe_{t \gi d}^{(i)} k_t - \left( I - g_{t \gi d}^{(i)} k_t k_t^{\top} \right) w_{t-1}^{(i)} \\
    & = k_t \left( g_{t \gi d}^{(i)} k_t^{\top} w_{t-1}^{(i)} + \beta_{t \gi d} r_{t \gi d}^{(i)} \pe_{t \gi d}^{(i)} \right). \quad \quad \blacksquare
\end{align*}

Because eq.~\eqref{eq:linearized} is a linear recurrence with identity-minus-rank-one transitions, each Newton iteration can be solved efficiently using the chunkwise algorithm. Note, however, that this is a \emph{quasi}-Newton iteration because these Jacobians are only approximate: they ignore the cross mixture component interactions coming from the denominator of the responsibilities in \cref{eq:responsibility}. However, \citet{gonzalez2024parallelizing} proves that quasi-Newton iterations are still guaranteed to converge globally, \emph{regardless} of the form of the Jacobian approximation. 

A limitation of quasi-Newton iterations, however, is that their backwards pass is only approximate \citep{gonzalez_thesis}. We can surmount this by running fixed-point iterations in the backwards pass as well; or, alternatively, by modifying the \method recurrence so that the Jacobians above are \emph{exact} (for example, by using "unnormalized" responsibilities in \cref{eq:responsibility}, i.e. removing the denominators, which is the only source of cross mixture component interaction). 

We have attempted these approaches: in configurations inspired by the language modeling pretraining settings (batch size of 16, sequence length $T=4096$, 8 heads, $D=64$, and $\GI=4$ mixture components), these various approaches converge in around 10 Newton iterations with synthetic inputs. However, despite this rapid convergence in Newton iterations, we so far did not optimize the parallel Triton kernels efficiently, so we were effectively the same speed as our sequential Triton kernel. Future work should be able to further accelerate such chunkwise parallel approaches for nonlinear recurrences.

\subsection{Chunkwise Parallel Algorithm}

To maximize hardware utilization, we implement the linearized recurrence
using a chunkwise parallel algorithm following~\citet{yang2024parallelizing}.
We try to follow the notation in \citet{yang2024deltanetblog2}.

In particular, we divide the sequence length into chunks of size $C$. We use $[n]$ to indicate boundary states and $\square_{[n]}^{c} = \square_{nC + c}$ to denote states within a chunk.
Within each Newton iteration~$i$, the computation proceeds in three steps.

\paragraph{1.\ Intra-chunk precomputation.}
For all timesteps~$t$ in parallel, we compute the Jacobian
coefficients~$g^{(i)}_{t \gi d}$ and effective inputs~$p^{(i)}_{t \gi d}$, and form
the strictly lower triangular dependency matrix
$\Lambda^{(i)}_{[n],\gi} \in \mathbb{R}^{C \times C}$ with entries
\begin{align}
    \bigl(\Lambda^{(i)}_{[n],\gi}\bigr)_{s,t}
    = \begin{cases}
        -g^{(i),s}_{[n],\gi}\,(k^s_{[n]})^\top k^t_{[n]}
            & \text{if } s > t, \\
        0   & \text{otherwise}.
    \end{cases}
    \label{eq:lambda}
\end{align}

\paragraph{2.\ Boundary state resolution.}
For each chunk~$n$, we compute the aggregate transition and history matrices via forward substitution through the unit lower triangular system $(I - \Lambda^{(i)}_{[n],\gi})$:
\begin{align}
    M^{(i)}_{[n],\gi} &= (I - \Lambda^{(i)}_{[n],\gi})^{-1}\,\operatorname{diag}(g^{(i)}_{[n],\gi})\,K_{[n]}, \label{eq:aggregate_transition} \\
    H^{(i)}_{[n],\gi} &= (I - \Lambda^{(i)}_{[n],\gi})^{-1}\,\operatorname{vec}(p^{(i),1}_{[n],\gi},\ldots,p^{(i),C}_{[n],\gi}).\label{eq:aggregate_history}
\end{align}
These aggregates allow the boundary states to be updated sequentially across chunks:
\begin{align}
    w^{(i+1)}_{[n+1],\gi} = w^{(i+1)}_{[n],\gi} + K^\top_{[n]} \bigl(H^{(i)}_{[n],\gi} - M^{(i)}_{[n],\gi}\,w^{(i+1)}_{[n],\gi}\bigr). \label{eq:boundary}
\end{align}

\paragraph{3.\ Parallel output generation.}
Given the boundary states, the outputs within each chunk are computed in parallel:
\begin{align}
    O^{(i+1)}_{[n],\gi} = K_{[n]}\,w^{(i+1)}_{[n],\gi} + \bigl(K_{[n]} K^\top_{[n]} \odot L\bigr) \bigl(H^{(i)}_{[n],\gi} - M^{(i)}_{[n],\gi}\,w^{(i+1)}_{[n],\gi}\bigr), \label{eq:parallel_output}
\end{align}
where $L \in \mathbb{R}^{C \times C}$ is a strictly lower triangular causal mask ensuring autoregressive validity.

\clearpage
\section{Experimental Details}\label{appendix:exp}

For all experiments, we heavily utilized architecture implementations in \citet{yang2024fla}.

\subsection{Approximating Softmax Attention}

\begin{figure}[!t]
\centering
\includegraphics[width=\textwidth]{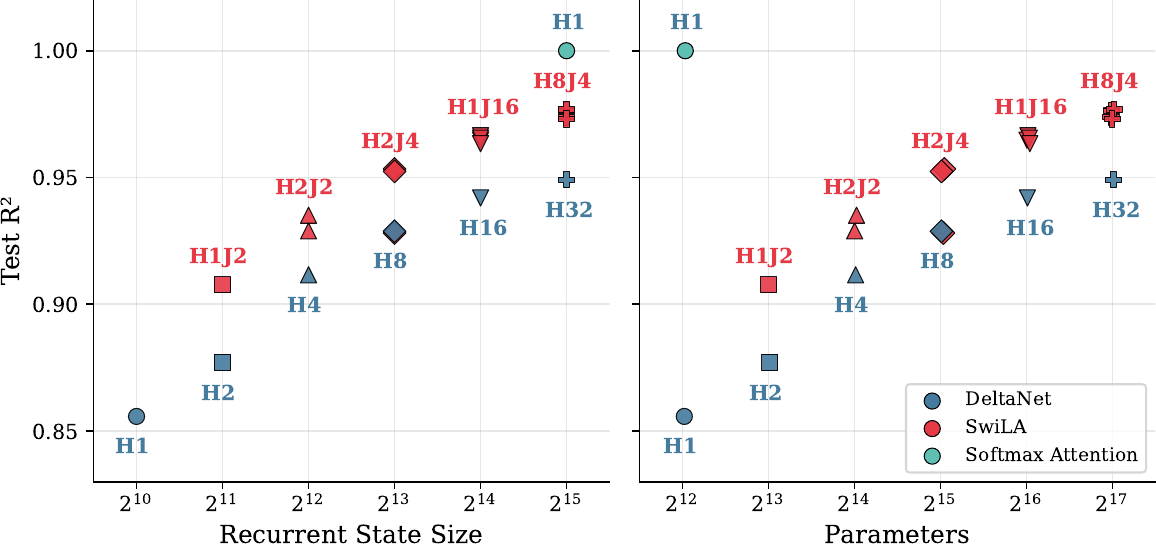}
\caption{\textbf{Approximating \Softmax Attention.}
Test $R^2$ vs.\ recurrent state size (left) and parameter count (right) for \method and DeltaNet on the synthetic attention regression task.
The dashed line indicates trainable \softmax attention.
Labels mark the best \method configuration (\texttt{H}$=$heads, \texttt{\GI}$=$mixtures) at each budget.
At every state budget, \method better approximates the attention mapping.}
\label{fig:regression_all}
\end{figure}

Each model receives input sequences $x \in \mathbb{R}^{T \times D}$, where $T=512$ and $D=32$, and must approximate the output of causal softmax self-attention with fixed, ground-truth projections $\theta_q, \theta_k, \theta_v$. 
Models receive only $x$ and must learn their own $Q$/$K$/$V$ projections.

\paragraph{Data Generation}
At each sequence position, a cluster index is sampled uniformly from $C = 32$ clusters. 
The input is $x_t = \mathrm{L2normalize}(\mu_c + 0.5\,\epsilon_t)$, where $\mu_c \sim \mathcal{N}(0, I_{32})$ are fixed cluster centers and $\epsilon_t \sim \mathcal{N}(0, I_{32})$. 
The query, key, and value projections $\theta_q, \theta_k, \theta_v \in \mathbb{R}^{32 \times 32}$ are random matrices, with each entry sampled from a standard normal distribution. 
The attention temperature is $\tau = \sqrt{D}/2$, and the sequence length is $T = 512$. 
We generate 65{,}536 training, 2{,}048 validation, and 2{,}048 test sequences.

\paragraph{Models.}
 
\begin{itemize}[leftmargin=*]
    \item \textbf{MHA.} Standard causal multi-head softmax attention with learned $W_q, W_k, W_v$ (bias enabled) and output projection $W_o$ (no bias). Scaling factor $1/\sqrt{d_k}$.
 
    \item \textbf{DeltaNet.} Learned projections with SiLU activation followed by L2 normalization on $Q$/$K$. SiLU activation on $V$. Scalar $\beta$ per head via a linear layer $\mathbb{R}^{d_{\mathrm{in}}} \to \mathbb{R}^{H}$ followed by sigmoid. RMSNorm on the output, then an output projection.
 
    \item \textbf{\pmethod.} Same $Q$/$K$/$V$ projection scheme as DeltaNet (SiLU + L2 normalization on $Q$/$K$, SiLU on $V$). Separate $\beta$ per head, output dimension, and mixture component. Routing logits are computed via SwiGLU projections (inner dimension set to 256) with softmax activation. RMSNorm on the output, then an output projection. MI auxiliary loss coefficient $\lambda = 0.001$. We do not employ input-dependent noise injection.
\end{itemize}

\subsection{Synthetic Benchmarks}

\paragraph{Context-Dependent Associative Recall (CDAR)}

Context-Dependent Associative Recall (CDAR) is a synthetic sequence modeling task that extends Multi-Query Associative Recall (MQAR) from \citet{zoology2024} to multiple contexts. Each input sequence has length 1024 and consists of two zones: a \emph{context zone} containing $C = 4$ distinct contexts, each defining 32 unique key--value pairs (128 total associations), followed by a \emph{query zone} in which the model must retrieve the correct value for a given (context, key) pair. The vocabulary size is 8192. Query positions within the query zone are sampled according to a power-law distribution $p(i) \propto i^{\,a-1}$ with exponent $a = 0.01$, following the Zoology benchmark setup \citep{zoology2024}.

We generate 100{,}000 training examples and 3{,}000 test examples. 
All queries for each context are grouped contiguously, context tokens are not repeated within query groups, and non-query positions are filled with zeros. 
Training uses a batch size of 64.

All models use a GPT-NeoX-style pre-norm block structure following the Zoology framework \citep{zoology2024}: each block consists of a sequence mixer followed by a state mixer (MLP, hidden multiplier 4), with LayerNorm, residual connections, and dropout on each sub-layer. The sequence mixer is the model-specific layer (Temporal \pmethod, DeltaNet, Softmax Attention, or MoM), while the state mixer is shared across all models. All models use a short convolution (kernel size 2) before the sequence mixer.

We detail the hyperparameters that differ across models:
 
\begin{itemize}[leftmargin=*]
    \item \textbf{Temporal \pmethod.} $d_{\mathrm{model}} \in \{32, 64, 128\}$, $n_{\mathrm{layers}} = 1$, $H = 1$ head, $K = 4$ mixture components, and key/value expansion factors $e_k = e_v = 0.5$. We used SwiGLU projections for $\pi$ and $\beta$, with inner dimensions 256 and 32, respectively, and input-dependent noise injection. We set the MI auxiliary loss coefficient $\lambda = 0.0$.

    \item \textbf{\pmethod.} $d_{\mathrm{model}} \in \{32, 64, 128, 256\}$, $n_{\mathrm{layers}} \in \{1, 2\}$, $H = 1$ head, $K = 4$ mixture components, and key/value expansion factors $e_k = e_v = 0.5$. We used SwiGLU projections for $\pi$ and $\beta$, with inner dimensions 256 and 32, respectively, and input-dependent noise injection. We set the MI auxiliary loss coefficient $\lambda = 0.0$.
 
    \item \textbf{DeltaNet.} $d_{\mathrm{model}} \in \{32, 64, 128, 256\}$, $n_{\mathrm{layers}} \in \{1, 2\}$, $H = 4$ heads, $e_k = e_v = 2.0$.
 
    \item \textbf{Softmax Attention.} $d_{\mathrm{model}} \in \{32, 64, 128, 256\}$, $n_{\mathrm{layers}} \in \{1, 2\}$, $H = 4$ heads, RoPE positional encoding with $\theta = 10{,}000$.
 
    \item \textbf{DeltaNet-MoM.} $d_{\mathrm{model}} \in \{32, 64, 128, 256\}$, $n_{\mathrm{layers}} \in \{1, 2\}$, $M = 4$ memories with top-$k = 2$ routing, $H = 1$ head, $e_v = 1.0$.
\end{itemize}

\begin{figure}[!t]
\centering
\includegraphics[width=0.6\textwidth]{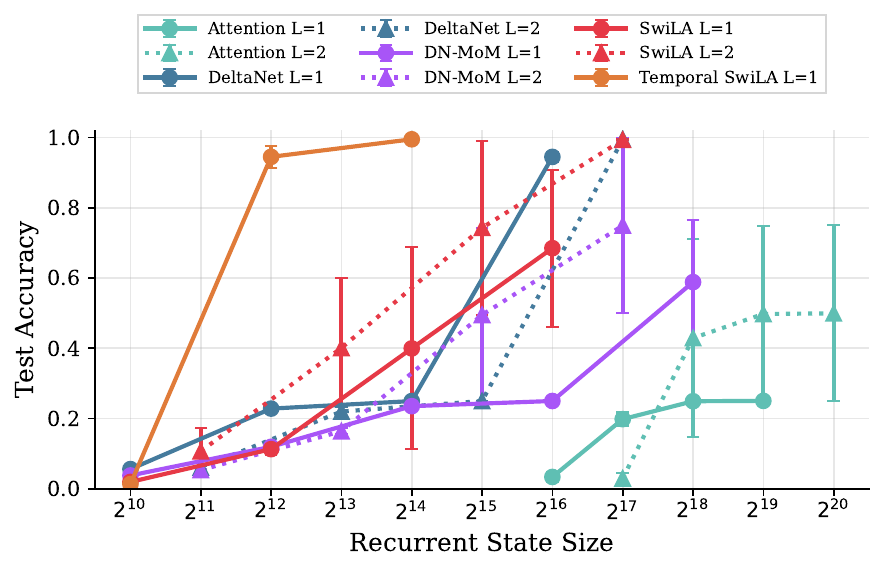}
\caption{\textbf{Context-Depedent Associative Recall.} CDAR with 4 contexts and 32 key-value pairs per context: test accuracy vs. state size.}
\label{fig:cdar_all}
\end{figure}

All models are trained with AdamW and a grid search over four learning rates log-spaced from $3.16 \times 10^{-4}$ to $10^{-2}$, i.e.\ $\{3.16 \times 10^{-4},\; 10^{-3},\; 3.16 \times 10^{-3},\; 10^{-2}\}$. All models are trained for 50 epochs. 
For each configuration and learning rate, we run three random seeds.
We report the maximum accuracy across the learning-rate sweep and three random seeds in \Cref{fig:cdar} because learning on these algorithmic tasks typically exhibits sharp, step-like transitions, and our primary objective is to assess whether an architecture can learn the target algorithm rather than its optimization speed or robustness.
For completeness, we report the mean test accuracy across learning rates in \Cref{fig:cdar_all}, which yields the same qualitative conclusions, with the largest improvements obtained by Temporal \pmethod.

\paragraph{RegBench: In-Context Language Learning}

All models have 8 layers, $d_{\text{model}} = 256$, MLP inner dimension $1{,}024$ ($4 \times d_{\text{model}}$), embedding dropout 0.1, and no residual dropout. 
\pmethod, Temporal \pmethod, DN-MoM, and DeltaNet all prepend a short convolution of width 4. 
We compare: \textbf{\method} (2 heads, 2 mixtures, $\texttt{expand\_k} = \texttt{expand\_v} = 0.5$); \textbf{Temporal \method} (same as \method with a temporal gate); \textbf{DN-MoM} (1 head, 4 memories, top-2 routing, $d_{\text{head}} = 64$, $\texttt{expand\_v} = 1.0$, DeltaNet backend); \textbf{DeltaNet} (4 heads, $\texttt{expand\_k} = \texttt{expand\_v} = 1.0$); and \textbf{Transformer} (4 heads with RoPE). The recurrent state size is matched across \pmethod, DN-MoM, and DeltaNet at $16{,}384 = 4 \times 64 \times 64$ per layer.
For both \method and Temporal \pmethod, we do not use input-dependent noise injection and set the MI loss coefficient to 0.005.
 
We use AdamW with learning rate $2.5 \times 10^{-4}$ and weight decay $0.1$. 
The learning rate follows a cosine schedule with 10\% linear warmup (from $10^{-6}$) and minimum learning rate $2.5 \times 10^{-5}$. 
Batch size is 32 and models are trained for 200 epochs. We report test accuracy using the checkpoint with the lowest validation loss.

\subsection{Pretraining Language Models}

All models are pretrained on a subset of the FineWeb-Edu dataset \citep{penedo2024fineweb} (\texttt{sample-100BT} subset) using the 32K-Llama2 tokenizer. Training sequences have length $4{,}096$.

All models are trained with identical optimization hyperparameters chosen following \citet{yang2024parallelizing}. We use the AdamW optimizer with a learning rate of $3 \times 10^{-4}$ and weight decay of $0.01$. The learning rate follows a cosine decay schedule with 954 warmup steps and a minimum learning rate ratio of $0.1$ (i.e., final learning rate $3 \times 10^{-5}$). Gradient norms are clipped to $1.0$. Training runs for 28{,}610 steps with a global batch size of $128$ sequences, totaling approximately 15B tokens. Training is conducted on a node of 8 NVIDIA H100 GPUs.

All thirteen models---Transformer++ \citep{touvron2023llama}, DeltaNet \citep{yang2024parallelizing}, GDN \citep{yang2025gdn}, two MoM variants (DeltaNet-MoM and GDN-MoM) \citep{du2026mom}, KDA \citep{team2025kimi}, FoX \citep{lin2025forgetting}, \pmethod, Temporal \pmethod, Gated \pmethod, Gated Temporal \pmethod, Hybrid GDN, and Hybrid Temporal \pmethod---share the same architecture skeleton: 24 layers and a hidden dimension of 1,024, targeting approximately 374M total parameters.
Feed-forward intermediate dimensions are adjusted slightly across architectures to match the total parameter count.
Following the default configurations in \citet{yang2025flame}, we use 16 heads with head dimension 64 for Transformer++. DeltaNet, GDN, KDA, and FoX use 8 heads with head dimensions of 128. 

\pmethod, Temporal \pmethod, Gated \pmethod, and Gated Temporal \method use 8 heads with 4 mixtures per head and 64-dimensional key and value projections. 
This gives a recurrent matrix state of $8\times4\times64\times64=131{,}072$ scalars per layer, matching the $8\times128\times128=131{,}072$ state scalars used by DeltaNet, GDN, and KDA. 
For all \method variants, we disable input-dependent router and temporal gate noise. 
We set the load balancing loss coefficients to $0.001$ for \pmethod, $0.0005$ for Temporal \pmethod, $0.0005$ for Gated \pmethod, and $0.00025$ for Gated Temporal \pmethod. 
For Gated \method variants, we tie the precision across the output index, i.e., one gate per key dimension, as in KDA \citep{team2025kimi}.
Hybrid Temporal \method uses the same $0.0005$ coefficient as Temporal \pmethod.
Hybrid GDN and Hybrid Temporal \method contain six full causal-attention layers at layers 4, 8, 12, 16, 20, and 24, with the remaining 18 layers using GDN and Temporal \pmethod, respectively. 
The full-attention layers use 8 heads with head dimension 128.

For GDN-MoM, we use GDN for the recurrent memory updates and employ three routed memories with top-$k=2$, together with one shared memory. 
Each memory uses 8 heads with head dimension 64, giving a total main recurrent state size of $4\times8\times64\times64=131{,}072$ scalars per layer. 

For DeltaNet-MoM, we trained the following three architectural configurations and report the one with the best average performance on the commonsense benchmark (best configuration in \textbf{bold}):
\begin{enumerate}
    \item 374.1M parameters: 4 heads, head dimension 64, 7 routed memories with top-$k=4$, one shared memory, and an MLP intermediate dimension of 2,634.
    \item 374.2M parameters: 1 head, head dimension 128, 7 routed memories with top-$k=4$, one shared memory, and an MLP intermediate dimension of 3,412.
    \item \textbf{374.1M parameters: 2 heads, head dimension 128, 3 routed memories with top-$k=2$, one shared memory, and an MLP intermediate dimension of 3,326.}
\end{enumerate}
Each of the three configurations maintains 131,072 main recurrent-state scalars per layer. 
Following \citet{du2026mom}, we use shared memory and adjust the MLP intermediate dimension to match the parameter counts of the other models.

\newpage
\subsection{Throughput Comparison}
\label{app:throughput}

\paragraph{Training throughput and memory.}

\begin{figure}[!htbp]
    \centering
    \includegraphics[width=\textwidth]{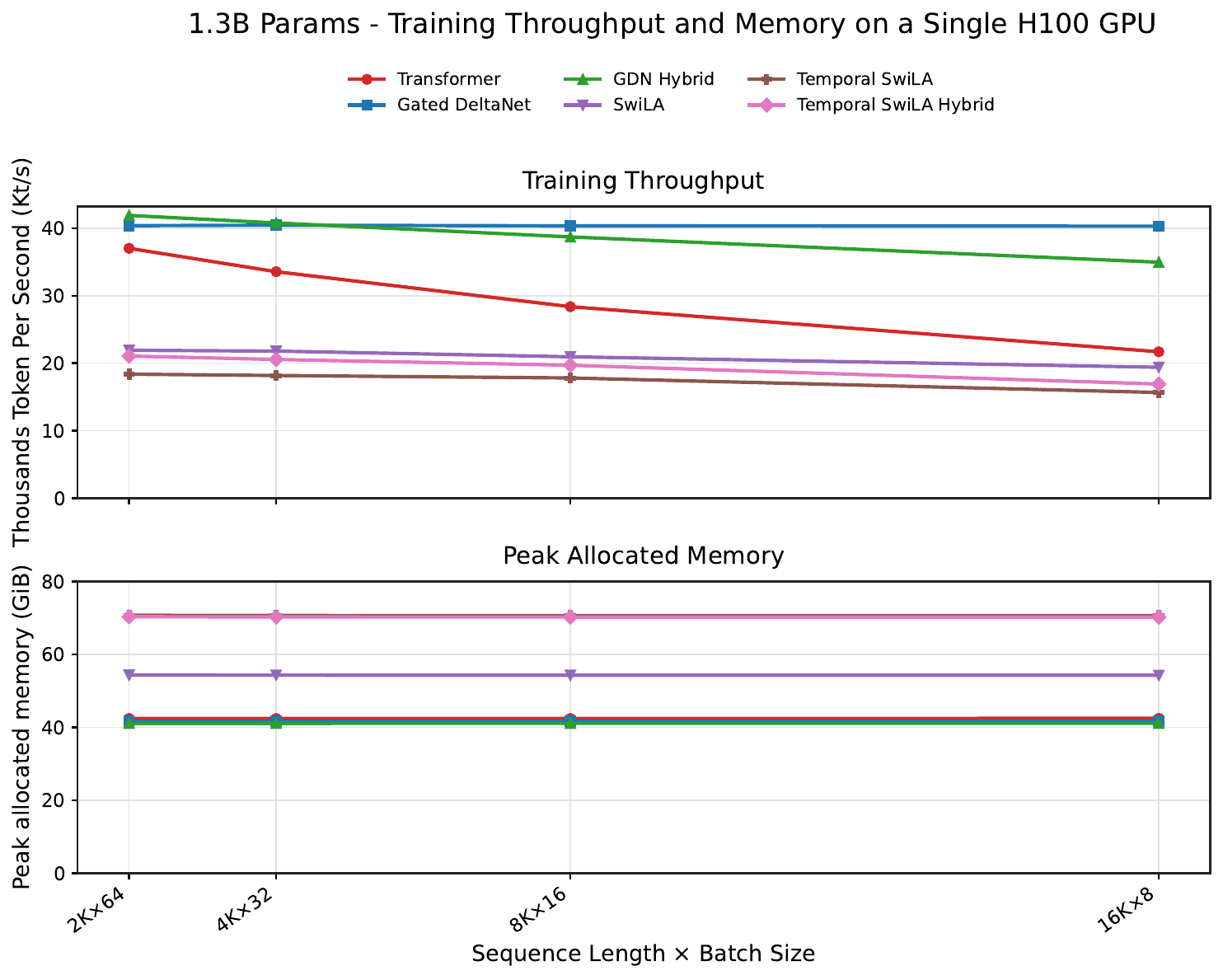}
    \caption{\textbf{Training throughput and peak memory of 1.3B models with 128K training tokens on a single H100 GPU (i.e., 1M training tokens on 8xH100 GPUs).} Following the training throughput experiment of \cite{yang2025gdn}, we fix the total number of training tokens while varying the sequence length and batch size.}
    \label{fig:train}
\end{figure}

We benchmark complete training updates on a single NVIDIA H100 for the six 1.3B models: Transformer, Gated DeltaNet (GDN), Hybrid GDN, \pmethod, Temporal \pmethod, and Hybrid Temporal \pmethod.
At sequence lengths 2,048, 4,096, 8,192, and 16,384 we use microbatch sizes 64, 32, 16, and 8, respectively, holding the workload fixed at 131,072 tokens per update.
A timed update comprises the forward pass and loss computation, the backward pass, global gradient-norm clipping, and an AdamW step.
For each configuration, three warm-up updates are followed by three timed updates with CUDA synchronization.
Throughput is 131,072 divided by the mean wall-clock update time, reported in thousands of tokens per second, and peak memory is PyTorch's maximum allocated CUDA memory after warm-up.

\paragraph{Generation throughput and memory.}

\begin{figure}[!htbp]
    \centering
    \includegraphics[width=\textwidth]{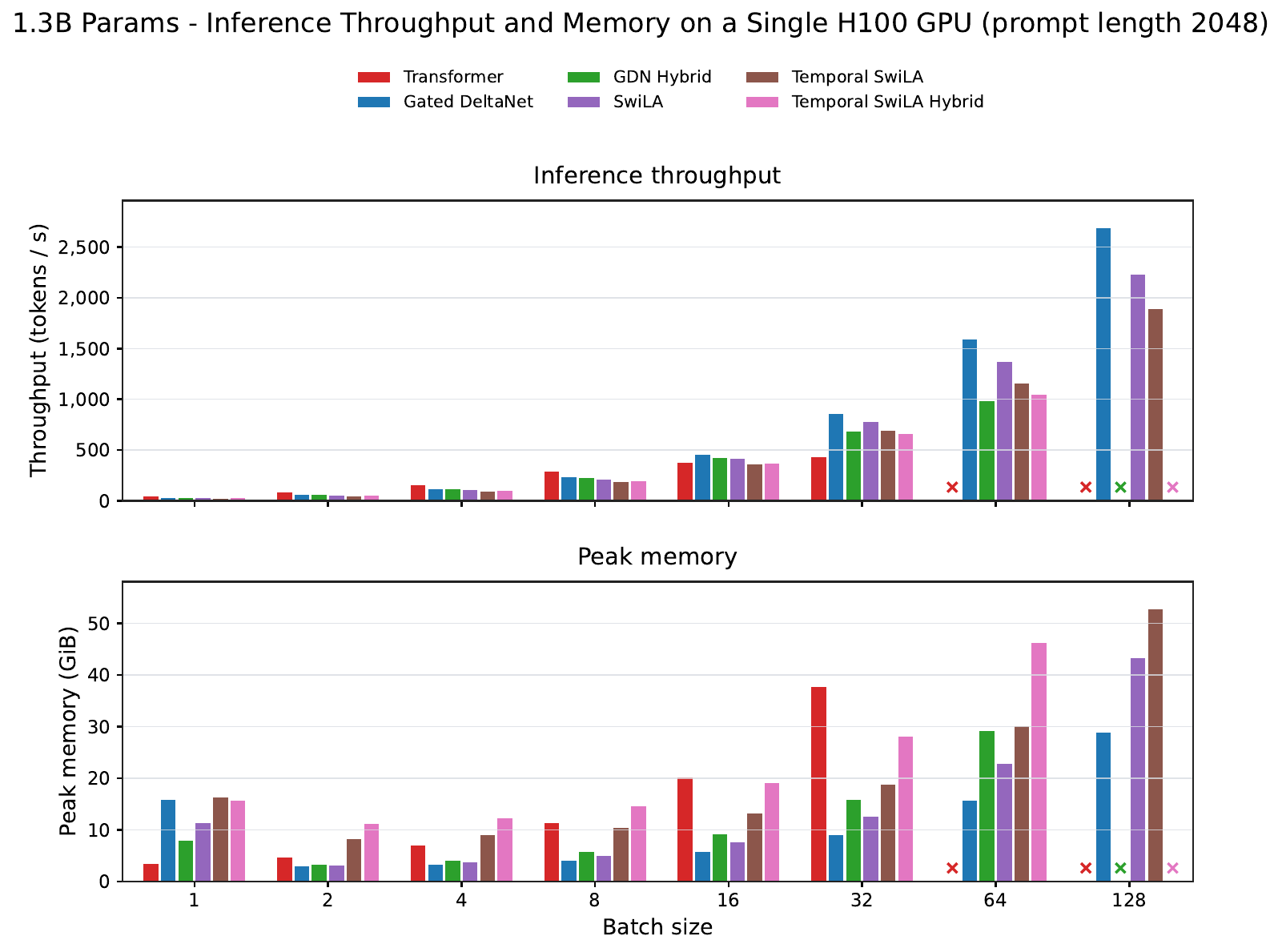}
    \caption{\textbf{Inference throughput and peak memory of 1.3B models on a single H100 GPU.} Following the inference throughput experiment of \citet{gu2023mamba}, we fix the prompt length to 2048 and vary the batch size. Throughput and peak memory are measured over the full inference pass, combining prefill and decode. An \textbf{\texttt{X}} marks configurations where the model runs out of memory.}
    \label{fig:inference}
\end{figure}

We measure end-to-end autoregressive generation for the same six models on a single NVIDIA H100.
Each input is a synthetic 2,048-token prompt extended by 128 generated tokens using greedy decoding.
We sweep batch sizes from 1 to 128 in powers of two.
For each model, batch sizes beyond the first out-of-memory failure are omitted.
One warm-up call is followed by three timed calls to the full generation routine, with CUDA synchronized around the timed loop.
Reported values are means over the three calls, where the wall-clock time of a generation call includes both prefill and decoding.
Peak memory is PyTorch's maximum allocated CUDA memory after warm-up.

\clearpage
\subsection{Ablation Study}
\label{app:routing}

\begin{figure}[!htbp]
    \centering
    \includegraphics[width=\textwidth]{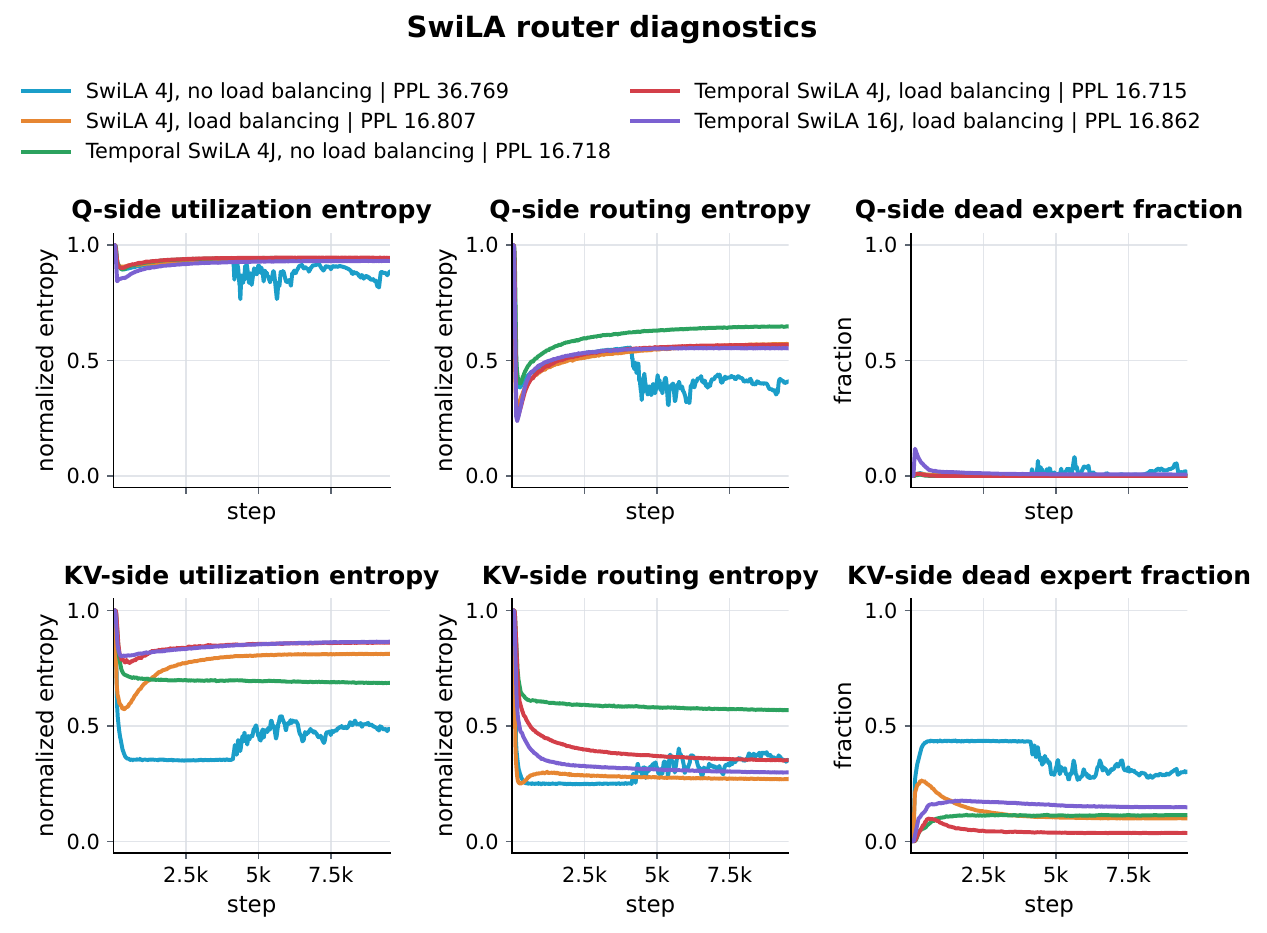}
    \caption{\textbf{Switching Linear Attention router diagnostics.} Utilization entropy, routing entropy, and dead expert fraction over the course of training, shown separately for the query-side responsibilities (top) and key-value-side responsibilities (bottom). Without load balancing, vanilla \method (4$J$) collapses on the KV side, with low utilization entropy and a dead expert fraction near 40\%; load balancing and temporal persistence each prevent this collapse.}
    \label{fig:router_diagnostics}
\end{figure}

We conducted ablation experiments that isolate the contribution of the number of mixture components, temporal persistence, and the load balancing loss. 
We trained variants of 115M \method on 5B tokens and evaluated test perplexity on 100M tokens, reported in the legend of Figure~\ref{fig:router_diagnostics}.
To analyze routing dynamics, we tracked three statistics during training, computed separately for the key-value-side responsibilities and the query-side responsibilities. 
\emph{Utilization entropy} measures the entropy of the marginal expert usage distribution, averaged over batch and sequence positions; it equals 1 when experts are used uniformly and is lower when usage is concentrated on a few experts. 
\emph{Routing entropy} measures the average per-token entropy of the routing distribution; it equals 1 when each token spreads its responsibility uniformly across all experts and approaches 0 when each token commits to a single expert. 
\emph{Dead expert fraction} counts the fraction of experts whose marginal usage falls below 10\% of uniform usage (i.e., below $0.1/J$).

Without load balancing, vanilla \method suffers significant expert collapse: KV-side utilization entropy drops sharply, and the dead expert fraction exceeds 40\%. 
Adding the load-balancing auxiliary loss resolves this, maintaining high utilization entropy and near-zero dead experts throughout training. Temporal \method is notably more robust, maintaining reasonable utilization even without load balancing. Across all configurations except for the vanilla \method without load balancing, routing entropy stabilizes in the 0.4–0.6 range, indicating moderately selective but not fully hard routing.

\end{document}